\documentclass[lettersize,journal]{IEEEtran}
\usepackage{amsmath,amsfonts}
\usepackage{algorithmic}
\usepackage{algorithm}
\usepackage{array}
\usepackage{textcomp}
\usepackage{stfloats}
\usepackage{url}
\usepackage{verbatim}
\usepackage{graphicx}
\usepackage{xcolor}
\usepackage{cite}
\usepackage{booktabs}
\usepackage{tabularx}
\usepackage{array}
\usepackage{subfigure}
\usepackage{subcaption}
\usepackage{color}
\usepackage{multirow}
\begin{document}

\title{RAG-based Crowdsourcing Task Decomposition via Masked Contrastive Learning with Prompts}

\author{
Jing Yang, Xiao Wang, ~\IEEEmembership{Senior Member, ~IEEE}, Yu Zhao, Yuhang Liu, Fei-Yue Wang, ~\IEEEmembership{Fellow,~IEEE}


\thanks{Jing Yang and Yuhang Liu is with The State Key Laboratory for Management and Control of Complex Systems, Institute of Automation, Chinese Academy of Sciences, Beijing 100190, China, and also with the School of Artificial Intelligence, University of Chinese Academy of Sciences, Beijing 100049, China (E-mail: yangjing2020@ia.ac.cn, liuyuhang2021@ia.ac.cn).}

\thanks{Xiao Wang is with the School of Artificial Intelligence, Anhui University, Hefei 266114, China and also with Qingdao Academy of Intelligent Industries, Qingdao 230031, China (E-mail: xiao.wang@ahu.edu.cn). }

\thanks{Yu Zhao is with National Key Laboratory of Information Systems Engineering, National University of Defense Technology (E-mail: yuzhao@nudt.edu.cn).}

\thanks{ Fei-Yue Wang is with Macau University of Science and Technology, Macao 999078, China, with Beijing Engineering Research Center of Intelligent Systems and Technology, Chinese Academy of Sciences, Beijing 100098, China and also with State Key Laboratory for Management and Control of Complex Systems, Chinese Academy of Sciences, Beijing 100190, China (e-mail: feiyue.wang@ia.ac.cn).} 

}

\markboth{Journal of \LaTeX\ Class Files,~Vol.~14, No.~8, August~2021}%
{Shell \MakeLowercase{\textit{et al.}}: A Sample Article Using IEEEtran.cls for IEEE Journals}


\maketitle

\begin{abstract}
Crowdsourcing is a critical technology in social manufacturing, which leverages an extensive and boundless reservoir of human resources to handle a wide array of complex tasks. The successful execution of these complex tasks relies on task decomposition (TD) and allocation, with the former being a prerequisite for the latter. Recently, pre-trained language models (PLMs)-based methods have garnered significant attention. However, they are constrained to handling straightforward common-sense tasks due to their inherent restrictions involving limited and difficult-to-update knowledge as well as the presence of “hallucinations”. To address these issues, we propose a retrieval-augmented generation-based crowdsourcing framework that reimagines TD as event detection from the perspective of natural language understanding. However, the existing detection methods fail to distinguish differences between event types and always depend on heuristic rules and external semantic analyzing tools. Therefore, we present a Prompt-Based Contrastive learning framework for TD (PBCT), which incorporates a prompt-based trigger detector to overcome dependence. Additionally, trigger-attentive sentinel and masked contrastive learning are introduced to provide varying attention to trigger and contextual features according to different event types. Experiment results demonstrate the competitiveness of our method in both supervised and zero-shot detection. A case study on printed circuit board manufacturing is showcased to validate its adaptability to unknown professional domains.

\end{abstract}


\section{Introduction}
\begin{figure}[!h]
    \centering  
    \includegraphics[width=1\linewidth]{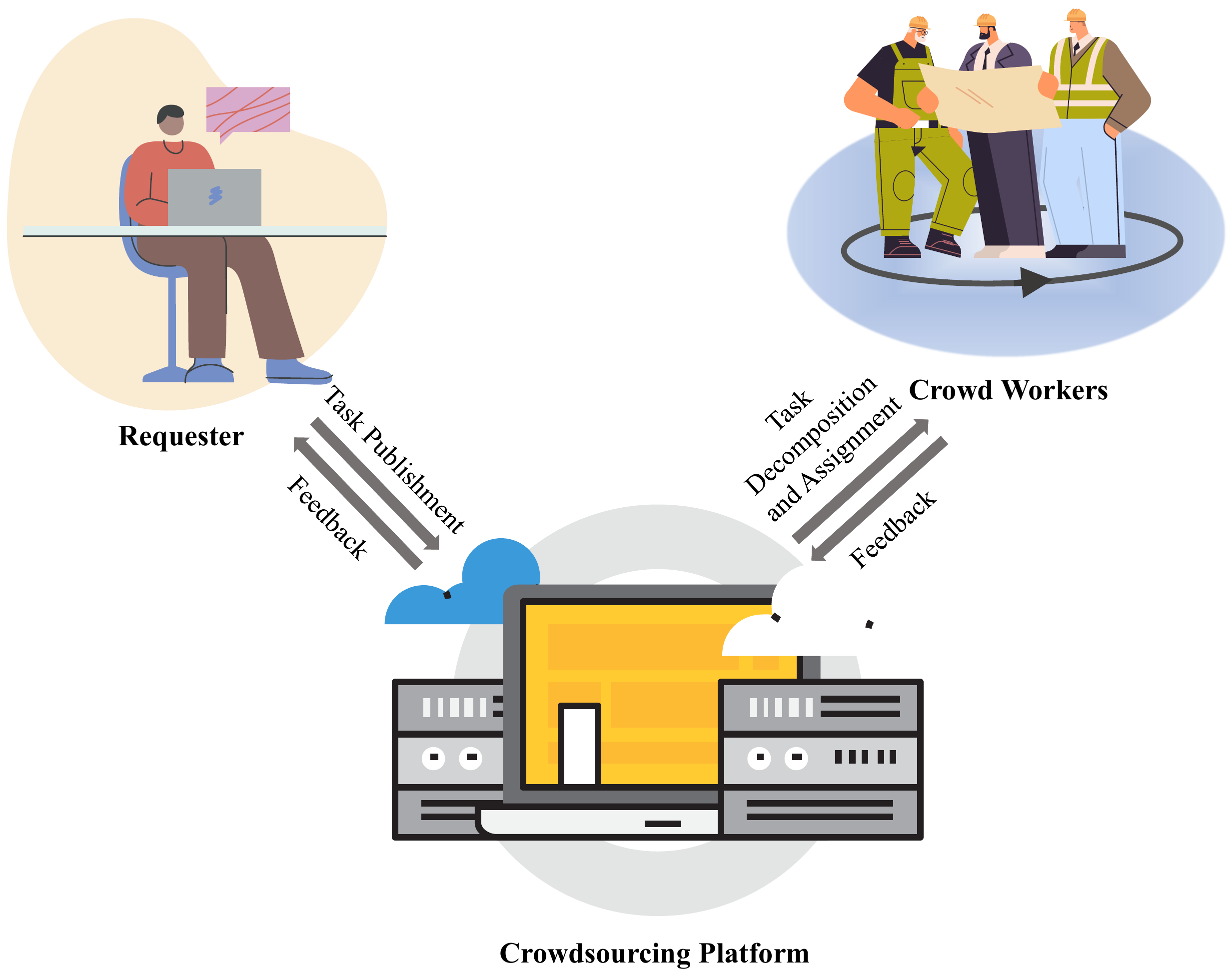}
    \caption{The workflow of crowdsouring.} 
    \label{fig: crowdsouring}
\end{figure}

Against the backdrop of increasingly diverse and personalized consumer demands, customized production is gaining significant attention and adoption as a pivotal production mode among numerous enterprises \cite{wang2024survey,yang2024GAI}. It not only caters to specific consumer needs and amplifies product differentiation for a competitive edge, but also curtails inventory wastage, enhances production flexibility, and fortifies customer loyalty \cite{Yang2022parallel,Yang2023DeFACT}. Crowdsourcing, epitomizing collective intelligence, as shown in Fig. \ref{fig: crowdsouring}, empowers businesses to gain profound insights into consumers' needs and preferences. Moreover, it even allows consumers to participate in the production process, leveraging an extensive, boundless reservoir of human resources to handle a wide array of tasks with varying levels of complexity \cite{xiong2017mind}. Therefore, it is crucial and essential to effectively apply crowdsourcing technology to customized production and thus promote the shift towards social manufacturing \cite{yang2023parallel,wang2023parallel}. 

\begin{figure}[!h]
    \centering  
    \includegraphics[width=1\linewidth]{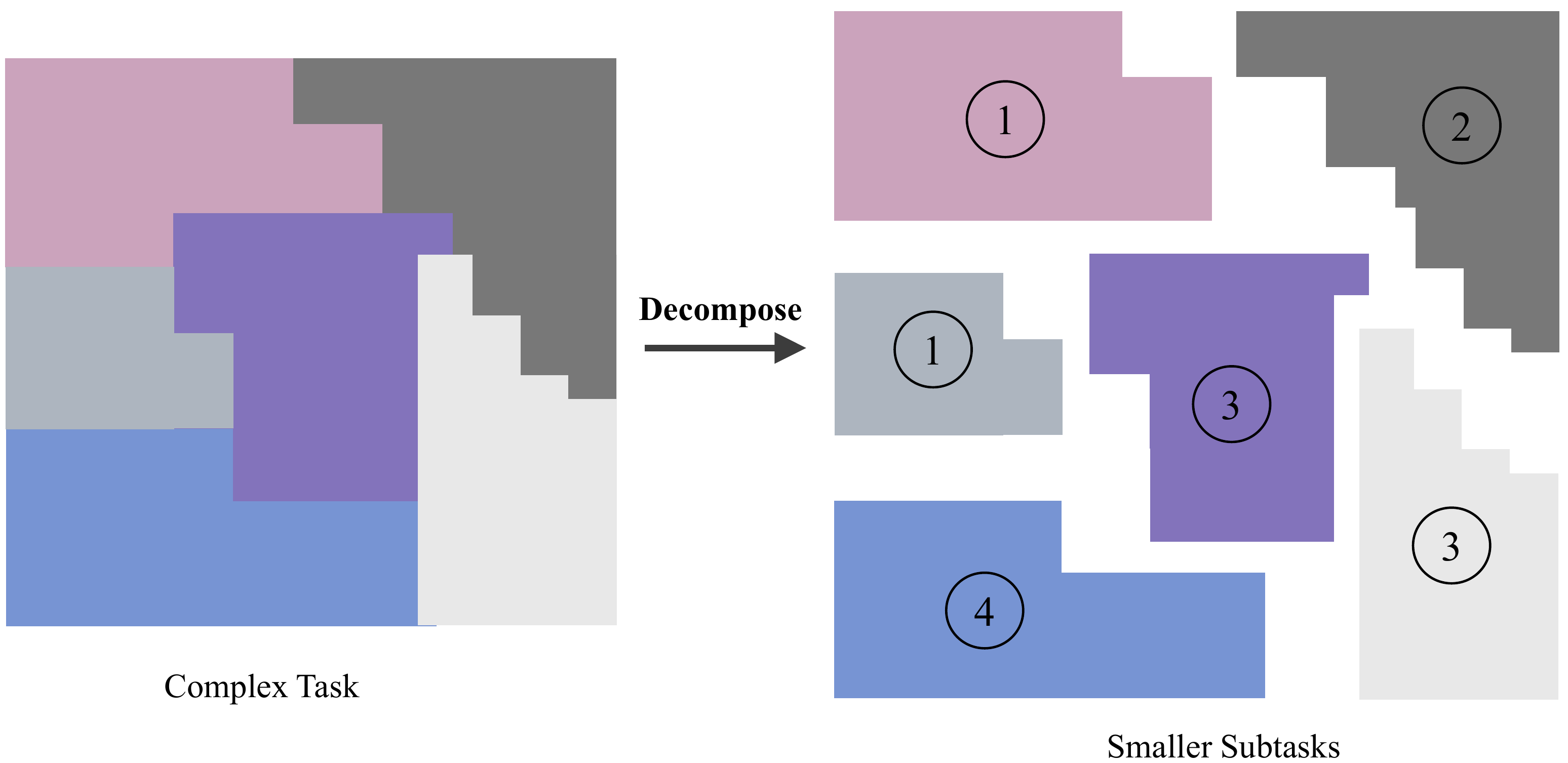}
    \caption{The illustration of Task decomposition. The numbers represents the order in which subtasks are carried out.} 
    \label{fig: deomposition}
\end{figure}

Requesters and crowd workers are organized and coordinated through a crowdsourcing platform, which acts as a mediator between them \cite{niu2017review}. Once a task is posted by a requester, the crowdsourcing platform allocates the task to a group of proficient crowd workers, with the goal of efficiently finishing the task within the shortest possible timeframe. Obviously, success depends on the ability to decompose complex tasks into smaller subtasks \cite{jiang2014efficient}, as illustrated in Fig. \ref{fig: deomposition}, each of which is low in complexity and requires minimal cognitive effort to be completed by an individual. These subtasks often feature interdependencies over time, and when orchestrated sequentially, they culminate in an event sequence delineating the complex task. However, prevalent TD methods often focus on specific scenarios, relying on prior knowledge of the complex task to design appropriate rules or optimization algorithms for its breakdown \cite{rizk2019cooperative,hu2014dynamic,li2010multi,shiarlis2018taco}. These approaches have significant limitations as they may not be easily adaptable to other scenarios and rely heavily on human expertise. Consequently, deploying them on crowdsourcing platforms for decomposing a variety of tasks proves challenging. 

\begin{figure*}[!h]
    \centering  
    \includegraphics[width=1\linewidth]{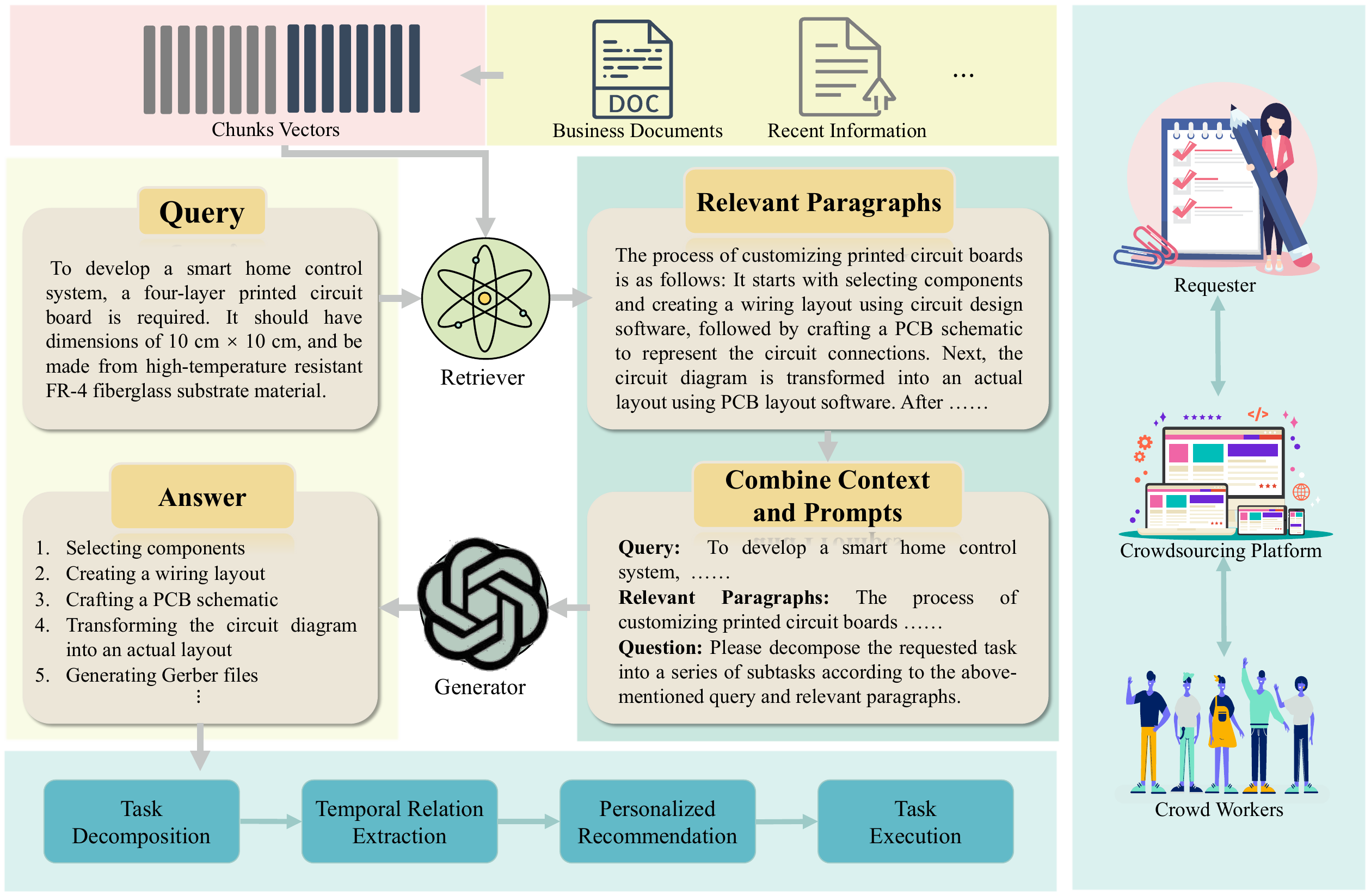}
    \caption{RAG-based crowdsourcing} 
    \label{fig: RAGCS}
\end{figure*}

Recently, pre-trained language models (PLMs) like BERT\cite{devlin2018bert}, ChatGPT\cite{wu2023brief}, and LLaMA\cite{touvron2023llama} have demonstrated the ability to store substantial world knowledge similar to large knowledge graphs during their pre-training phase. These models have showcased exceptional performance across a variety of tasks, such as event extraction \cite{gao2023exploring}, temporal relation extraction \cite{yuan2023zero} and personalized recommendation \cite{li2023preliminary}. Given that requesters’ tasks are typically described in natural language, a straightforward idea is to treat task decomposition as a natural language understanding and information extraction task using PLMs. Several attempts have been made to stimulate the inferential capabilities of PLMs and activate their stored knowledge via prompts for TD \cite{jansen2020visually,sakaguchi2021proscript,kannan2023smart}. However, the current PLMs-based TD is primarily constrained to handling common sense tasks due to their limited and hard-to-update knowledge. Another prevalent challenge involves the generation of inaccurate information, known as “hallucinations”, particularly when queries surpass the PLM's training data. 

One promising approach to address these issues is to integrate retrieval-augmented generation (RAG) \cite{lewis2020retrieval} into the crowdsourcing systems, which incorporates retrieved external data into the TD process, thereby improving the model's capability to accurately decompose diverse tasks. Consequently, based on the pertinent retrieved data for the task's execution steps, we extract or detect the triggers and related arguments of subtasks (which can be also considered as events) for each step. It is evident that TD tasks can be converted into event extraction or detection tasks, an area where significant effort has already been invested. However, the existing detection methods might have only learned lexical patterns \cite{liu2020does}, i.e., word-to-trigger mapping, resulting in skewed performance as they fail to distinguish differences between event types. Additionally, with the constant emergence of new tasks or events without annotated samples in the open world, it is imperative and necessary to deal with zero-shot detection tasks. Therefore, the main contributions of this paper are as follows. 
\begin{itemize}
    \item We propose an RAG-based crowdsourcing framework that addresses challenges related to hallucinations and limited knowledge through external retrieval. By this means, TD tasks are transformed into event detection tasks, contributing to the utilization of mature detection techniques.
    
    \item We present a prompt-based contrastive learning framework for TD, named PBCT, which incorporates prompt learning for trigger detection. It breaks away from dependency on heuristic rules and external semantic analyzing tools.  
     
    \item We introduce a trigger-attentive sentinel and masked contrastive learning to provide varying attention to trigger features and contextual features according to different event types. This method brings similar samples closer and pushes away samples of different classes, thereby improving the representation learning of unseen types under a zero-shot setting. 
    \item We conduct a series of experiments on two datasets: ACE 2005 and FewEvent, and the results demonstrate that our method achieves competitive performance in both supervised and zero-shot detection. Additionally, we offer a case study on personalized printed circuit board (PCB) manufacturing to validate the effectiveness of our approach for TD in RAG-based crowdsourcing.    
\end{itemize}

The subsequent sections of this paper are structured as follows. Section \uppercase\expandafter{\romannumeral2} presents the RAG-based crowdsourcing framework and its operational workflow, followed by a review of related work for TD in Section \uppercase\expandafter{\romannumeral3}. Section \uppercase\expandafter{\romannumeral4} introduces our problem statement and relevant notations. In Section \uppercase\expandafter{\romannumeral5}, a comprehensive overview of our model design is presented, along with a detailed explanation of each component. Section \uppercase\expandafter{\romannumeral6} gives and discusses a series of experimental results. Finally, concluding remarks and future work are drawn in Section \uppercase\expandafter{\romannumeral7}.

\section{RAG-based Crowdsourcing}
Crowdsourcing is a goal-directed and self-organizing collective behavior. The interaction between requesters and crowd workers is crucial for the efficiency and quality of crowdsourced task completion. Therefore, it is important to accurately grasp the requirements of requesters and convert them into a series of subtasks allocated to crowd workers. The emergence of PLMs and RAG makes it possible to automate their interactions. To enhance PLMs’ inference performance and equip them with the capability to handle complex, diverse and open-world tasks, we reformulate all computational tasks, such as TD, temporal relation extraction, and personalized recommendation,  within crowdsourcing systems into prompt engineering tasks via RAG modes. Fig. \ref{fig: RAGCS} illustrates the framework of RAG-based crowdsourcing, which interfaces with an external knowledge repository to augment and rectify the knowledge stored within PLMs, especially professional and up-to-date knowledge.  During the preparatory phase, the corpus in the repository is segmented into discrete chunks, which are embedded into vector representations as their indices through an encoder. 

In terms of the task-processing phase, we take task decomposition as an example in Fig. \ref{fig: RAGCS} to explain the operational workflow of the framework. After a requester initiates a query, the same encoder as the preparatory phase is utilized to transform the input into a vector representation. Subsequently, the similarity between the query vector and index vector is computed to obtain the most relevant chunks via a retriever. Finally, the query, along with the relevant paragraphs retrieved and a task-specific prompt, are combined as input for the generator to produce answers.  Based on the TD results, we sequentially extract the temporal dependencies of subtasks and recommend corresponding subtasks to the competent crowd workers, following the same procedure. Once the tasks are assigned, the crowd workers would independently or collaboratively strive to accomplish the tasks, driven by incentive mechanisms \cite{wang2023smart,li2023dao}.

In this paper, our focus is on enhancing the performance of the generator used to identify a series of subtasks based on a pertinent paragraph describing a complex task. Therefore, we assume that the retriever can access critical texts regarding the steps and measures relevant to task execution, which is expected to be the focal point of our future research.

\section{Task Decomposition}
TD is the first step in crowdsourcing task automation, for which many efforts have been made. Some approaches necessitate designers to manually break down intricate tasks into simpler executable ones \cite{dorigo2013swarmanoid,kiener2010towards,kittur2011crowdforge}. Especially, to reduce the barrier to entry for workflow design, Turkomatic \cite{kulkarni2012collaboratively} is introduced as a tool that engages crowd workers to adaptively decompose and solve tasks. Besides, there are semi-automated modes as the prevailing trend. They are often task-specific \cite{xie2021multi} or object-specific \cite{motes2020multi,shiarlis2018taco,botelho1999m+}, heavily relying on prior knowledge of task structures and subtask repositories. Their focus is to design TD rules and evaluation metrics tailored to specific scenarios, and then refine various TD strategies through intelligent optimization algorithms. Tong et al.\cite{tong2018slade} propose an optimal priority queue-based approach to decompose large-scale crowdsourcing tasks that consist of thousands or millions of independent atomic subtasks in both homogeneous and heterogeneous scenarios. However, this method struggles to address complex and interdependent tasks. Motes et al. \cite{motes2020multi} concentrate on multi-robot transportation tasks and integrate decomposition, allocation, and planning for these multiple decomposable tasks to seek the optimal solution through technologies like conflict-based search. Zhang et al. \cite{zhang2016task} leverage an integrated numerical design structure matrix and an adaptive genetic algorithm to divide product development tasks into different groups based on their correlation degrees.  For complex control tasks, researchers consider they can be solved by multiple sub-polices corresponding to subtasks, so learning from demonstrations is utilized to build sequential decision processes \cite{shiarlis2018taco,cobo2012automatic}. In software crowdsourcing, a general-purpose approach is presented to verify the complex task’s decomposition scheme and its functional specification \cite{shu2016verification}. Song et al. employ a design structure matrix and a product structure tree to model the interaction of the design process and thus decompose the design task for complex products \cite{song2021novel}. 
While these methods to a certain extent achieve automation in TD, none of them are generalizable across different task characteristics. 

Recently, the incorporation of PLMs into TD has garnered some attention. Sakaguchi et al. \cite{sakaguchi2021proscript} finetune PLMs to generate high-quality scripts including partial ordering events through two complementary tasks: edge prediction and script generation. Jansen et al.\cite{jansen2020visually} model visual semantic planning as a translation problem of converting natural language directives into detailed multi-step sequences of actions and finetune PLMs to generate a command sequence via prompts. SmartLLM \cite{kannan2023smart} employs Pythonic chain-of-thought prompts and demonstrations to guide the large PLMs in generating code for multi-robot task planning including TD within the context of smart homes. However, the capabilities of these methods are constrained within addressing prototype tasks in everyday scenarios that demand common-sense knowledge. It is non-trivial to the application of PLMs in dealing with more complex tasks, especially professional tasks. Additionally, it is challenging to enhance their decomposition performance due to hallucination issues. This paper aims at providing a unified solution for decomposing various types of tasks by introducing RAG to model TD as event extraction or detection tasks. 

\begin{figure}[!t]
    \centering  
    \includegraphics[width=1\linewidth]{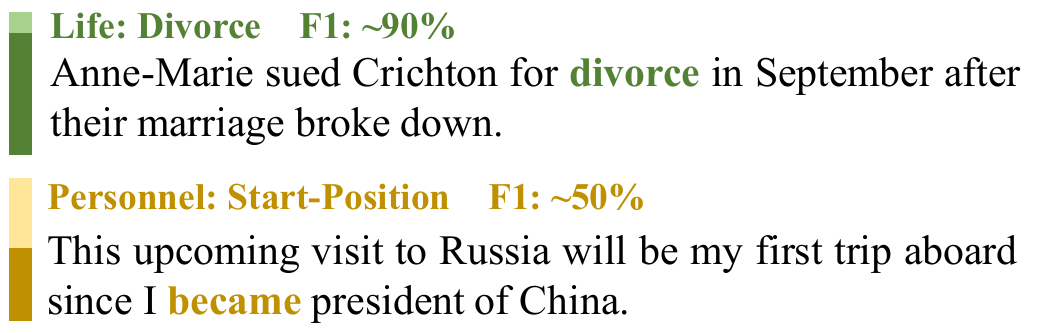}
    \caption{Two typical instances taken from the ACE 2005 benchmark.} 
    \label{fig: instance}
\end{figure}

\begin{figure*}[!t]
    \centering  
    \includegraphics[width=1\linewidth]{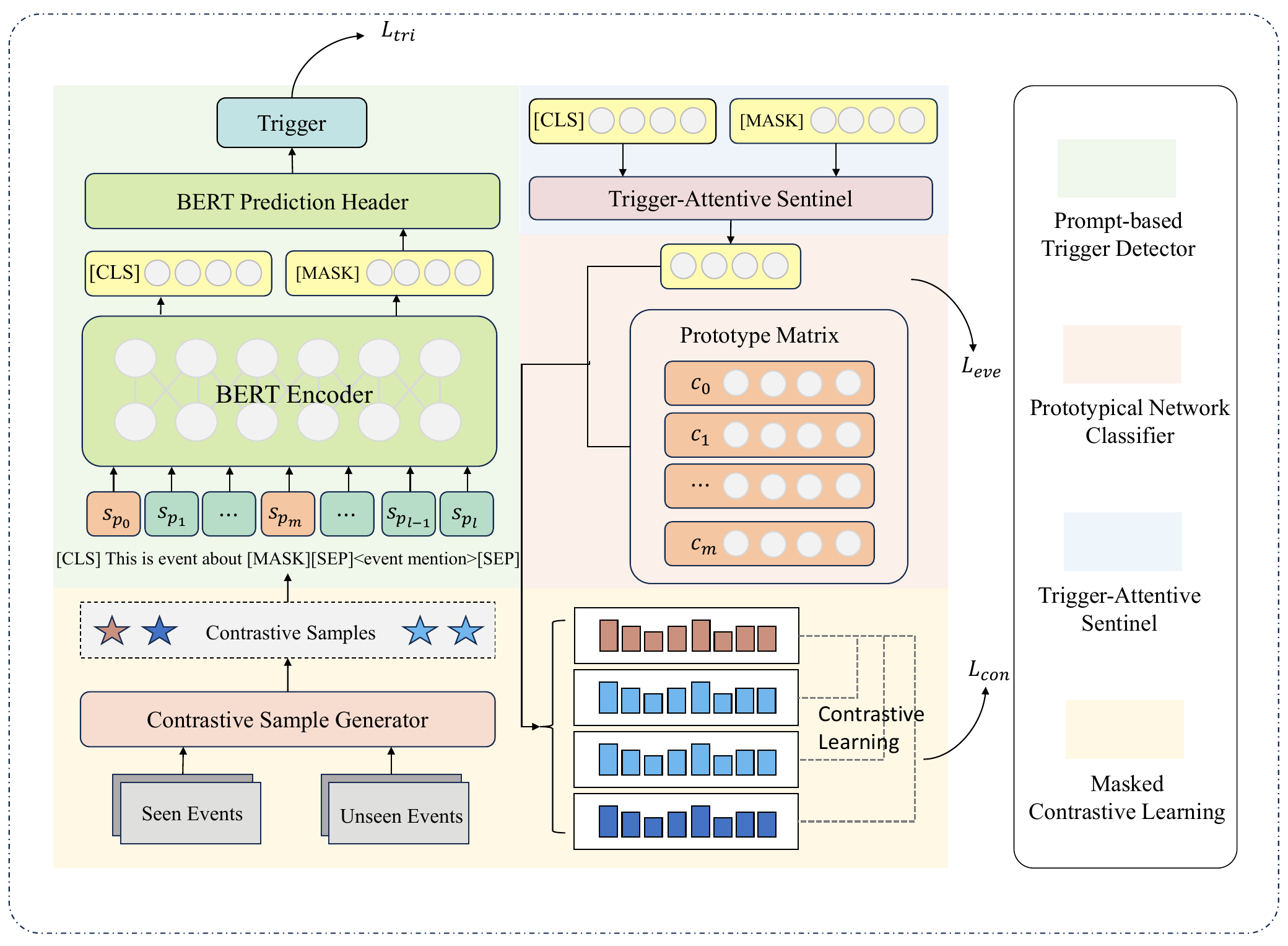}
    \caption{The overall architecture of PBCT, which consists of four modules: A prompt-based trigger detector, a prototypical network classifier, a trigger-attentive sentinel and masked contrastive learning.} 
    \label{fig: architecture}
\end{figure*}

Event detection is a relatively well-established field that focuses on identifying specific types of events (also known as different subtasks in TD), within unstructured text. Extensive research has been conducted in both supervised and few-shot/zero-shot settings. Despite considerable advancements in supervised training\cite{lin2020joint, wadden2019entity, liu2020event, du2020event, lu2021text2event, liu2022dynamic}, few-shot/zero-shot detection poses ongoing difficulties owing to the limited availability of prior knowledge and annotated examples, a common issue in the open world. Under zero-shot settings, most existing works rely on additional semantic analysis tools to supplement event knowledge or design heuristic rules to facilitate the detection of unseen events\cite{huang2017zero,huang2020semi,lyu2021zero,zhang2021zero,li2023glen}. Zhang et al.\cite{zhang2021zero} compute the similarity between triggers and arguments to the type's names to classify events with the help of external tools (e.g., Semantic Role Labeling) and corpora (e.g., New York Times). Huang and Ji \cite{huang2020semi} propose a semi-supervised vector quantized variational autoencoder framework that discovers candidate triggers with a heuristic approach and projects them into a particular type. However, these methods fail to recognize the distinction between event types and train a model that treats all categories equally, leading to imbalanced performance. Observations \cite{liu2020does,liu2022saliency} indicate that performance tends to be relatively lower on context-dependent texts, as models may primarily learn lexical patterns. For instance, as shown in Fig.\ref{fig: instance}, while DYGIE++ \cite{wadden2019entity} achieves a high F1 score of 90\% for the \textit{Life: Divorce} type, its performance drops to 50\% for the \textit{Personnel: Start-Position} type. What's more surprising is that the training set for \textit{Personnel: Start-Position} is eight times larger than that of \textit{Life: Divorce}. This suggests that, in the case of context-dependent texts, the implied semantics of triggers may be inadequate, resulting in an inability to classify events using lexical patterns. In response to this challenge, our approach focuses on adjusting attention between trigger features and contextual features by employing trigger-attentive sentinel and masked contrastive learning techniques.

\section{Problem Statement}
Given an event mention as an input $S=\lbrace s_0, s_1, \ldots, s_N\rbrace$, including one or more trigger words annotated with event types $y \in \mathcal T $, where $s_i$ is the $i$-th token and $N$ is the number of tokens. A pre-defined set of event types $\mathcal T$ consists of seen event types $\mathcal T_s$ and unseen event types $\mathcal T_u$, that is, $\mathcal T=\mathcal T_s \bigcup \mathcal T_u$, $\mathcal T_s \bigcap \mathcal T_u= \emptyset$. During the training phase, unseen event types are not used as sample labels to optimize the model. Additionally, the triggers of these samples remain invisible to the model. The goal of this task is to identify the triggers and subsequently classify events into appropriate types based on them and the corresponding sentences.

\section{Model Architecture}
In this section, we elaborate on our proposed PBCT model for TD from the perspective of event detection, with Fig. \ref{fig: architecture} illustrating its overall architecture, comprising four modules: a prompt-based trigger detector, a prototypical network classifier, a trigger-attentive sentinel, and masked contrastive learning. The prompt-based trigger detector leverages prompt learning to fully exploit the inference capability of PLMs for trigger identification, eliminating the need for additional semantic analysis tools or heuristic rule design. The prototypical network classifier trains a prototypical matrix for event types, which are initialized by label semantics and thus classify events through similarity calculation. The trigger-attentive sentinel balances attention between the trigger and contextual features by learning two weights. Masked contrastive learning is employed to enhance event representation learning, achieved by constructing two positive samples and one negative sample, along with their corresponding losses. Additionally, to take great advantage of all available information, it is also indispensable to incorporate supervised learning in the detection of seen events.

\subsection{Prompt-based Trigger Detector}
As the first step of event detection, a prompt-based trigger detector is introduced to acquire the contextual vectors of triggers. We construct a prompt template $PT=$ “This is an event about [MASK]. $\left\langle event \: mention\right\rangle$” to input a PLM for trigger prediction, where [MASK] refers to a trigger word to be predicted and $\left\langle event \: mention\right\rangle$ represents the description of an event. Before input, $PT$ is transformed into the input sequence $S_p=\lbrace s_{p_0}, s_{p_1}, \ldots, s_{p_m}, \ldots,  s_{p_{l-1}},s_{p_l}\rbrace$ via a tokenizer, where $s_{p_m}$ is the token at the [MASK] position and $l$ is the length of the tokenized $PT$. To achieve more accurate predictions, we fine-tune the model using the cross-entropy loss function $\mathcal L_{tri}$ for seen events and thereby obtain a word distribution $P_{m}^t(s_{p_i})=P_{m}^t(s_{p_m}=s_{p_i}|S_p)$ over words in the $\left\langle event \: mention\right\rangle$: 

\begin{equation}\label{eq1}
\mathcal L_{tri}=
\begin{cases}
-\sum_{i=8}^l P_{m}^{gt}(s_{p_i}) log P_{m}^t(s_{p_i})   & y\in \mathcal T_s  \\
0  & y\in \mathcal T_u  \\
\end{cases}
\end{equation}
where $P_{m}^{gt}(s_{p_i})=P_{m}^{gt}(s_{p_m}=s_{p_i}|S_p)$ is a ground-truth word distribution and $i=8$ is the index indicating the starting position of $\left\langle event \: mention\right\rangle$ in $PT$. 

\subsection{Prototypical Network Classifier}
After trigger detection, a prototypical network classifier is introduced to achieve event-type prediction.  Specifically, a prototype matrix $C=[\mathbf{c}_0, \mathbf{c}_1,\ldots,\mathbf{c}_m]^{T} \in \Bbb R^{m \times h}$ is built, with each row representing the prototype of an embedded event type, while $h$ denotes the embedding dimension of the PLM and $m=\lvert \mathcal T \rvert$ is the number of event types. 

To provide a solid starting point for the prototype network, we propose a semantic initialization approach to expedite model convergence and boost performance. Specifically, event-type labels are joined as $XL$ = “$y_0, y_1, \ldots, y_m$”.  Afterward, we utilize the PLM to encode the input sequence $XL$ consisting of type labels, thereby segmenting the output $\mathbf{H}_{XL}$ to acquire each type representation $\mathbf{h}_{ y_j}$:


\begin{equation}\label{eq1}
\mathbf{H}_{XL}=\mathit{PLM}(XL)
\end{equation}


\begin{equation}\label{eq1}
{\mathbf{h}_{ y_0},\mathbf{h}_{y_1}, \ldots,  \mathbf{h}_{y_m}}=\mathit{Segment}({\mathbf{H}_{XL}}).
\end{equation}

It is worth noting that, in this paper, we choose a PLM based on self-attention as their attention heads facilitate direct interactions between type words. This approach can take great advantage of event-type labels and cross-type interaction relations and offer important clues for type representation.

Finally, we derive the predicted probabilities $P(y=y_j|\mathbf{x})$ of the queried event description across each type by calculating the similarity between its event vector $\mathbf{x}$ and the prototype of the event type $\mathbf{c}_i$:

\begin{equation}\label{eq1}
P(y=y_j|\mathbf{x})=\frac{exp(-d(\mathbf{x},\mathbf{c}_i))}{\sum_{y_k\in \mathcal T}exp(-d(\mathbf{x},\mathbf{c}_k))}
\end{equation}
where $d(\cdot,\cdot)$ is set as Euclidean distance. 

Therefore, the event type with the high probability is chosen as the final prediction $\widehat{y}$:
\begin{equation}\label{eq1}
\widehat{y}=argmax_{y_j \in \mathcal T} P(y=y_j|\mathbf{x})
\end{equation}

\subsection{Trigger-Attentive Sentinel}

With each event type associated with multiple trigger words, there's a risk that relying solely on predicted trigger words as query points may cause samples of the same event type to be widely dispersed across hidden spaces. Additionally, some trigger embedding vectors may lack distinctive semantics, with the bulk of semantic information contained within contextual embedding vectors. Therefore, we introduce a trigger-attentive sentinel to learn the trade-off between trigger features and contextual features through the computation of two attention weights, $g_0$ and $g_1$:

\begin{equation}\label{eq1}
\begin{aligned}
g_0,g_1=\sigma(\mathbf{W}_g[\mathbf{E}_{[CLS]}\bigoplus \mathbf{E}_{[MASK]}]+\mathbf{b}_g) \\ s.t. \quad g_0+g_1=1
\end{aligned}
\end{equation}
where $\mathbf{E}_{[CLS]}$ and $\mathbf{E}_{[MASK]}$ refer to a contextual embedding vector and a trigger embedding vector, respectively, encoded by PLMs in the prompt-based trigger detector, $\mathbf{W}_g$ and $\mathbf{b}_g$ are learnable parameters of the feedforward network, and $\sigma$ denotes the softmax function. Afterward, a weighted summation of $\mathbf{E}_{[CLS]}$ and $\mathbf{E}_{[MASK]}$ is calculated to derive a queried vector $\mathbf{x}$ corresponding to an event mention:  
\begin{equation}\label{eq1}
\mathbf{x}=g_0\mathbf{E}_{[CLS]}+g_1\mathbf{E}_{[MASK]}
\end{equation}

This sentinel allows the model to dynamically adjust the combination of the two features based on trigger saliency, contributing to the final prediction. 

\subsection{Masked Contrastive Learning}
To better represent events under the zero-shot setting, especially unseen events, we design a masked contrastive learning mechanism by constructing several contrastive samples, i.e., two positive samples and one negative sample. Specifically, for positive sample One, we rephrase event mentions through back translation \cite{fadaee2018back} as original event triggers are kept. This can ensure the consistency of semantics, so the rephrased sample is regarded as a positive sample. For positive sample Two, we aim to reduce lexical bias and encourage the model to make predictions based on context alone. To achieve this, we delexicalize triggers by replacing them with placeholders “[MASK]” \cite{liu2020does}, so the masked sample can serve as a positive sample. In contrast, the selection of negative samples should be divided into two situations: for seen events, seen events with different labels or unseen events  are randomly selected, while for unseen events, negative samples are chosen from all seen events.

Therefore, the distances between the probability distributions of contrastive samples and the original sample are calculated using the Wasserstein distance as the distance function $\Bbb W_p$, denoted as $d_1, d_2$ and $d_3$, as follows: 

\begin{equation}\label{eq1}
d_i=\Bbb W_p(p_i,p_0), i\in \lbrace 1,2,3\rbrace
\end{equation}

\begin{equation}\label{eq1}
\Bbb W_p(p_i, p_j)= inf_{\gamma \in \prod(p_i,p_j)} \boldsymbol E_{(x,y) \sim \gamma}[c(x,y)] 
\end{equation}
where $p_0,p_1,p_2,p_3$ refers to event type probability distribution of the original, rephrased, masked and negative sample, $\prod(p_i,p_j)$ represents the set of all possible joint distributions combined by two distributions $p_i$ and $p_j$, $(x,y)$ is a sample drawn from a joint distribution $\gamma$, $c(x,y)$ denotes the cost function for transferring $x$ to $y$ and is typically set as the $L_p$ distance, $inf$ refers to the minimum expected cost.

The goal of contrastive learning is to bring positive samples closer and push negative samples farther away. However, certain negative samples start with a considerable distance from the original sample. Taking these distances into account may lead to a shift in learning focus, potentially leaving smaller distances unexpanded. Hence, it is necessary to establish a margin $m_1$ to define the negative samples that require pushing further away, as follows:

\begin{equation}\label{eq1}
\widehat{d}_3=max(0, m_1-d_3).
\end{equation}

\begin{figure}[!t]
    \centering  
    \includegraphics[width=1\linewidth]{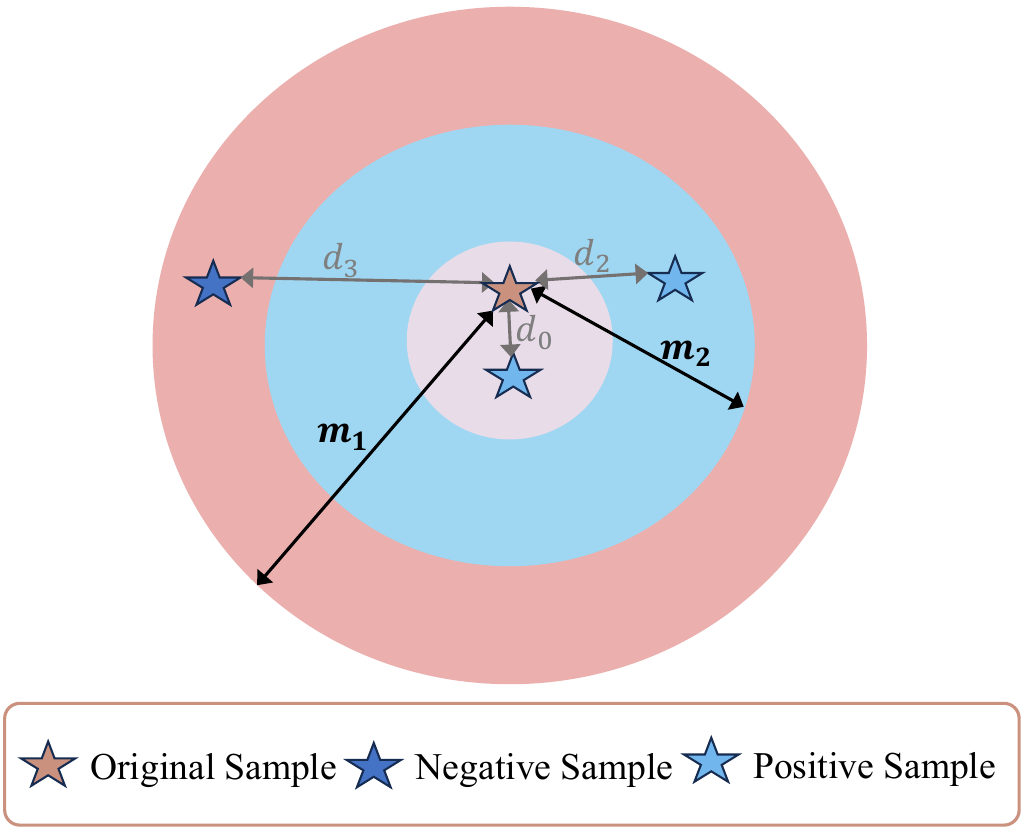}
    \caption{The schematic of masked contrastive learning.} 
    \label{fig: MCL}
\end{figure}

Moreover, for certain event mentions, the removal of trigger words has minimal impact on semantic expression, resulting in a relatively close distance between the masked sample and the original sample. However, there are instances where eliminating the trigger words render the expressed event type unrecognizable even to humans because the majority of the semantics of this event mention is implied by its trigger words. If these masked samples with larger distances are forcefully brought closer to the original sample, it could mislead the representation of other events, given the substantial overlap of tokens between masked event mentions and mentions of other event types. Therefore, similar to negative samples, we must design a margin $m_2$ to mask out masked samples that extend beyond it, as follows: 

\begin{equation}\label{eq1}
\widehat{d}_2=min(m_2, d_2).
\end{equation}

Fig. \ref{fig: MCL} depicts a schematic of the masked contrastive learning mechanism. The total contrastive loss is calculated as: 
\begin{equation}\label{eq1}
\mathcal L_{con}=d_1+\widehat{d}_2+\widehat{d}_3
\end{equation}

\begin{table*}[t]
  \centering
  \caption{Dataset split and statistics. The last two columns are the mean and standard deviation of samples by types.}
  \label{tab:DATA}
  \setlength{\tabcolsep}{0.5mm}{
  \resizebox{0.7\linewidth}{!}{
  \begin{tabular}{c|c|c|c|cc|cc|c|c}
    \toprule
    \multirow{2}{*}{Dataset}&\multirow{2}{*}{Split} &\multirow{2}{*}{Total Sample}  &\multirow{2}{*}{Total Type}& \multicolumn{2}{c|}{Seen} &\multicolumn{2}{c|}{Unseen}  &\multirow{2}{*}{Mean}&\multirow{2}{*}{Stdev} \\
    & & & & Sample& Type& Sample& Type &  &   \\
    \midrule
    \multirow{4}{*}{ACE 2005}& Total& 3805&  \multirow{4}{*}{33}& 2316 & \multirow{4}{*}{17}& 1489& \multirow{4}{*}{16}& 115.30& 21.60\\
    & Training &3043 & &1852& &1191& &92.21&17.26\\
    & Validation & 381 & &232& &149 & & 11.55 &2.17\\
    & Test &381 & &232& & 149 & & 11.55 &2.17\\
    \midrule
    \multirow{4}{*}{FewEvent}&Total&74330&\multirow{4}{*}{100}&40891&\multirow{4}{*}{50}&33439&\multirow{4}{*}{50}&743.30&74.52\\
    & Training &59463 & & 32712& &26751& &594.63&59.61\\ 
    & Validation & 7433 & & 4089& & 3344 & & 74.33 & 7.45\\
    & Test & 7434 & & 4090& & 3344 & & 74.33 & 7.46\\
    \bottomrule
  \end{tabular}
}}
\end{table*}

\subsection{Supervised Loss}
Since the labels of seen event samples are available to the model, supervised learning can be applied alongside contrastive learning, to optimize the first $\lvert \mathcal T_s \rvert$ rows of the prototype matrix corresponding to seen event types. To this end, the cross-entropy loss $\mathcal L_{eve}$ for event classification is calculated as: 

\begin{equation}\label{eq1}
\mathcal L_{eve}=
\begin{cases}
-\sum_{i=0}^{\lvert \mathcal T_s \rvert-1}P^{gt}(y=y_i|\mathbf{x}) log P(y=y_i|\mathbf{x})  & y\in \mathcal T_s  \\
0  & y\in \mathcal T_u  \\
\end{cases}
\end{equation}
where $P^{gt}(y=y_i|\mathbf{x})$ is a ground-truth event type distribution. 

Therefore, the total loss is the summation of the supervised loss and the contrastive loss:  

\begin{equation}\label{eq1}
\mathcal L=\mathcal L_{eve}+ \mathcal L_{tri}+ \lambda*\mathcal L_{con} 
\end{equation}
where $\lambda$ is a trade-off hyper-parameter between the two types of losses. 

\section{Experiments}
\subsection{Datasets and Baselines}
We take ACE 2005 \footnote{https://catalog.ldc.upenn.edu/LDC2006T06} \cite{ahn2006stages} and FewEvent \footnote{https://github.com/231sm/Low\_Resource\_KBP} \cite{deng2020meta} as the benchmark datasets. ACE 2005 stands out as the most widely-used dataset for event detection, drawing from a diverse array of news sources spanning 6 distinct categories. It includes annotations for 8 event types and 33 more finely-grained subtypes. FewEvent currently stands as the largest few-shot dataset for event detection, covering 19 event types, which are further subdivided into 100 event subtypes. To ensure a balanced distribution of samples between seen and unseen types, we arrange the event types in descending order according to their respective sample counts, where odd-positioned event types are considered as seen types and even-positioned ones as unseen types. Additionally, these datasets are split into training, validation and test sets in proportions of 80\%, 10\% and 10\%. According to the statistics, while FewEvent has more samples and types than ACE 2005, it is facing a more conspicuous sampling bias issue. More details about the two datasets and their splits are shown in TABLE \ref{tab:DATA}.

\begin{table*}[t]
  \centering
  \caption{Comparison of various detection methods on ACE 2005 and FewEvent datasets. The best results are bolded and the second-best results are underlined. }
  \label{tab:result}
  \setlength{\tabcolsep}{0.5mm}{
  \resizebox{1\linewidth}{!}{
  \begin{tabular}{c|ccccccc|ccccccc}
    \toprule
    \multirow{2}{*}{Model}& \multicolumn{7}{c|}{ACE 2005}& \multicolumn{7}{c}{FewEvent}\\ 
     & F1-Seen & F1-Unseen & NMI & FM & P & R & F1 & F1-Seen & F1-Unseen & NMI & FM & P & R & F1\\  
    \midrule
    SCCL & 0.5936 & 0.2893 & 0.2607 & 0.2185 & 0.5543 & 0.4987 & 0.4746 & 0.7892 & 0.3140 & 0.2832 & 0.2855 & 0.7374 & 0.6032 & 0.6250\\ 
    SS-VQ-VAE & 0.7233 & 0.2812 & 0.1202 & \textbf{0.4096} & 0.5205 & 0.6116 & 0.5504 & 0.9232 & 0.4344 & 0.1710 & 0.5754 & 0.7226 & 0.7608 & 0.7033\\
    \midrule
    BERT-MCL-Base & 0.4108 & 0.2765 & 0.3312 & 0.1682 & 0.4702 & 0.3753 & 0.3583 & 0.8681 & 0.5285 & \textbf{0.4722} & 0.6149 & 0.7368 & 0.5075 & 0.7153\\
    BERT-MCL-Large & 0.4686 & 0.3359 & \textbf{0.3641} & 0.2168 & 0.5222 & 0.4252 & 0.4167 & 0.8710 & 0.4611 & 0.4119 & 0.5561 & 0.7167 & 0.6691 & 0.6866 \\
    \midrule
    PBCT-Base & \textbf{0.8153} & \textbf{0.4327} & \underline{0.3633} & \underline{0.3934} & \underline{0.6790} & \textbf{0.6693} & \textbf{0.6657} & \underline{0.9389} & \underline{0.5609} & 0.4536 & \underline{0.7471} & \underline{0.7957} & \underline{0.7998} & \underline{0.7689}\\
    PBCT-Large & \underline{0.8017} & \underline{0.3527} & 0.3027 & 0.3361 & \textbf{0.6823} & \underline{0.6194} & \underline{0.6261} & \textbf{0.9454} & \textbf{0.5634} & \underline{0.4575} & \textbf{0.7514} & \textbf{0. 8055} & \textbf{0.8056} & \textbf{0.7736}\\
    \bottomrule
  \end{tabular}
}}
\end{table*}

To showcase its superiority, our methods are compared against the following baselines:
\begin{itemize}
    \item \textbf{SCCL}\cite{zhang2021supporting} It is a state-of-the-art model designed for unsupervised text clustering, which leverages instance-wise contrastive learning to support unsupervised clustering. Instead of using the [CLS] token as the representation for event mentions, we employ the contextual vector of identical candidate trigger words as SS-VQ-VAE, tailored to fit the event detection task.
    \item \textbf{SS-VQ-VAE}\cite{huang2020semi} It is a semi-supervised vector quantized variational autoencoder approach that associates the sense of each word with OntoNotes \cite{hovy2006ontonotes}, consequently considering all mapped noun and verb concepts as candidate triggers.
    \item \textbf{BERT-MCL} We fine-tune BERT-Base\footnote{https://huggingface.co/google-bert/bert-base-uncased} and BERT-Large\footnote{https://huggingface.co/google-bert/bert-large-uncased} with our proposed masked contrastive learning. KNN algorithm for seen event detection and K-means algorithm for unseen event detection are utilized once event encoding is obtained. 
\end{itemize}

\begin{table*}[t]
  \centering
  \caption{Results of ablation studies. The best results are bolded.}
  \label{tab:ablation}
  \setlength{\tabcolsep}{0.5mm}{
  \resizebox{1\linewidth}{!}{
  \begin{tabular}{l|ccccccc|ccccccc}
    \toprule
    \multirow{2}{*}{Model}& \multicolumn{7}{c|}{ACE 2005}& \multicolumn{7}{c}{FewEvent}\\ 
     & F1-Seen & F1-Unseen & NMI & FM & P & R & F1 & F1-Seen & F1-Unseen & NMI & FM & P & R & F1\\  
    \midrule
    PBCT-Base& \underline{0.8153} & \textbf{0.4327} &\textbf{0.3633} & 0.3934 & \textbf{0.6790} & \textbf{0.6693} & \textbf{0.6657} & \textbf{0.9389} & \textbf{0.5609} & \textbf{0.4536} & \textbf{0.7471} & 0.7957 & \textbf{0.7998} & \textbf{0.7689}\\
    -wo Atten. & 0.7998 & 0.3301 & 0.1774 & \textbf{0.4963} & 0.6257 & 0.6562 & 0.6161 & 0.9351 & 0.5541 & 0.4322 & 0.7337 & 0.8019 & 0.7963 & 0.7637\\
    -wo SemanInit. & 0.7820 & 0.3274& 0.1800 & 0.4935 & 0.6138 & 0.6536 & 0.6042 & 0.9306 & 0.5559 & 0.4338 & 0.7356 & 0.7918 & 0.7920 & 0.7621\\
    -wo MaskContra. & \textbf{0.8418} & 0.3545& 0.3377& 0.3564 & 0.6772 & 0.6614 & 0.6512 & 0.9347 & 0.5519 & 0.4283 & 0.7308 & \textbf{0.8063} & 0.7957 & 0.7625\\
    \bottomrule
  \end{tabular}
}}
\end{table*}

\subsection{Training Settings and Evaluation Metrics}
We code and implement our PBCT model by using PyTorch on a Linux server with four GPU Tesla V100. BERT-Base and BERT-Large are chosen as the PLM for the prompt-based trigger detector. Therefore, the total parameters of the executed PBCT are 109.59M and 335.28M, respectively. Argos Translate\footnote{https://www.argosopentech.com} is employed for back translation, with Chinese designated as the intermediary language. For the ACE 2005 dataset, we employ the Adam optimizer with a learning rate of $1e-6$ and a weight decay of $1e-6$ for the parameters of BERT, alongside a learning rate of $1e-3$ for the remaining parameters. Conversely, for the FewEvent dataset, we adjust the learning rate to $1e-5$ for the BERT parameters and $1e-2$ for the remaining parameters. Our evaluation task is composed of two parts: supervised event detection for seen events and zero-shot event detection for unseen events, mirroring real-world scenarios in open environments. Our model directly outputs the predicted labels of seen events and maps its output into the predicted labels of unseen events by the Hungarian Algorithm \cite{zhang2022zero}. We use the F1 score as the same metric for the two parts, i.e., F1-Seen and F1-Unseen and especially leverage Normalized Mutual Info (NMI) and Fowlkes Mallows (FM) \cite{huang2020semi,zhang2021supporting} to assess the clustering performance for unseen events. Additionally, we opt for weighted precision, recall, and F1 to evaluate the overall performance in both supervised and zero-shot settings, denoted as P, R, F1. We tune the PBCT’s hyperparameters according to F1 scores on each validation set. The margins $m_1$ and $m_2$ of masked contrastive learning are set to 1 and 0.6. The trade-off hyperparameter $\lambda$ between losses is assigned to 0.7. 

\subsection{Results and analysis}
We compare the test results of our PBCT model with those of baselines for event detection on the ACE 2005 and FewEvent datasets, as shown in Table \ref{tab:result}. Our model surpasses the state-of-the-art model on most metrics and demonstrates comparable performance on remaining metrics. BERT-MCL achieves competitive results for unseen events, thus highlighting the effectiveness of masked contrastive learning in zero-shot training scenarios. SS-VQ-VAE's superior performance for seen events over SCCL and BERT-MCL underscores the indispensability and necessity of supervised learning. PBCT outperforms BERT-MCL across most metrics in detecting unseen events, which demonstrates the prototype network's superior ability to learn representations for each type. The superior performance of PBCT over SS-VQ-VAE suggests that a prompt-based trigger detector holds its own against heuristic rules. PBCT-Base generally exhibits better performance than PBCT-Large on the ACE 2005 dataset, but it underperforms PBCT-Large on the FewEvent dataset. We believe this is because the smaller data scale of ACE 2005 led to overfitting or underfitting in the PBCT-Large model with a larger parameter size.

PBCT achieves remarkable enhancements, with a 9.2\% improvement in F1-Seen and 9.68\% in F1-Unseen over the best baseline on the ACE 2005 dataset, along with gains of 2.22\% in F1-Seen and 3.49\% in F1-Unseen on the FewEvent dataset. This showcases that PBCT has a slightly greater advantage in zero-shot unseen event detection compared to other baselines. Our model also achieves improvements of  12.8\% in P, 5.77\% in R and 11.53\% in F1 on the ACE 2005 dataset as well as 6.81\% in P, 4.48\% in R and 5.83\% in F1 on the FewEvent dataset, compared to the state-of-the-art results. It is observed that our model's performance improvement on the ACE 2005 dataset surpasses its improvement on the FewEvent dataset, suggesting an edge in scenarios with limited available data.

\begin{figure}[!t]
    \centering  
    \includegraphics[width=1\linewidth]{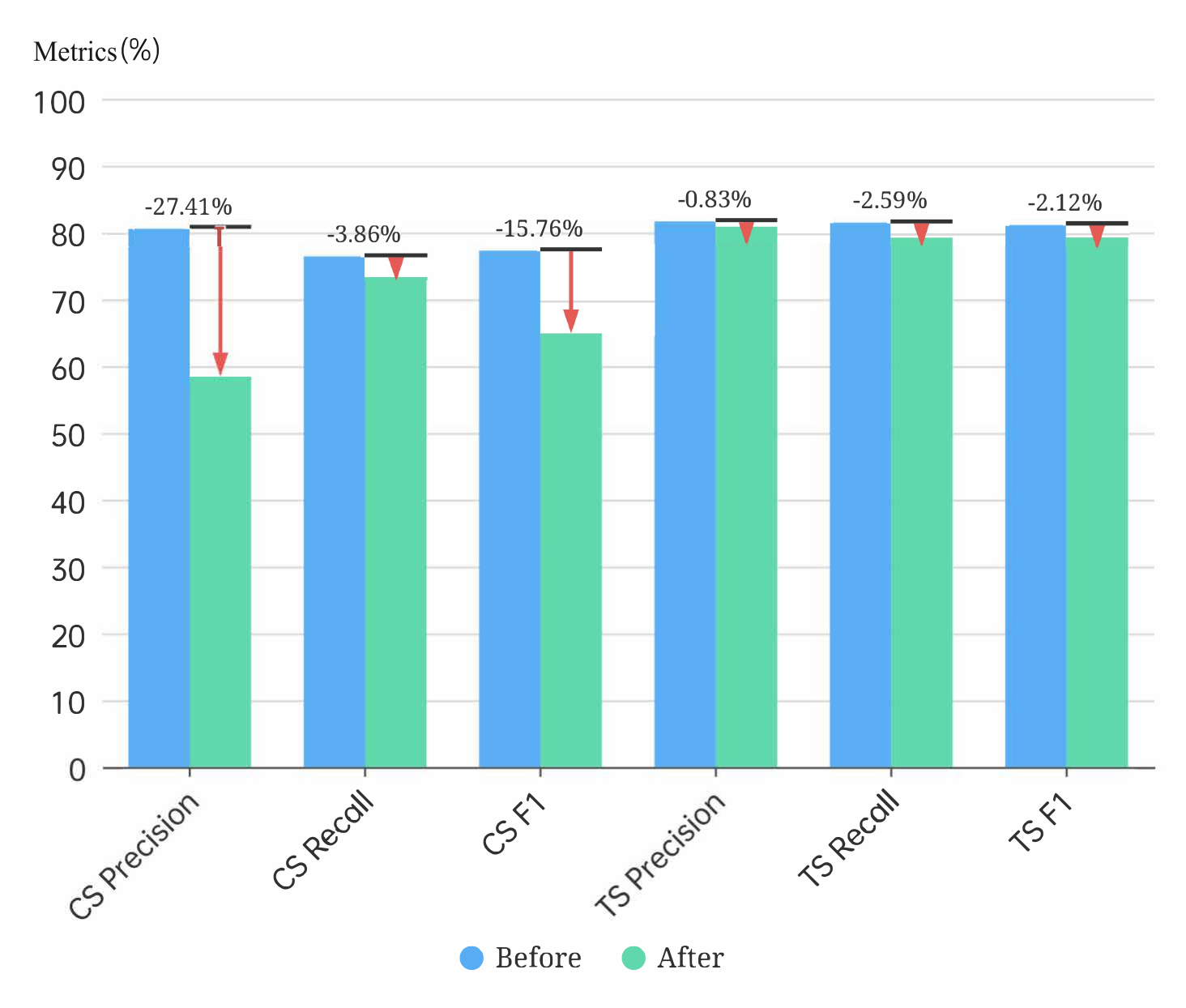}
    \caption{The performance change in event detection of trigger-salient (TS) and context-salient (CS) types before and after removing masked contrastive learning and trigger-attentive sentinel.} 
    \label{fig: CS}
\end{figure}

\begin{figure*}[htbp]
\centering 
\subfigure[$m_1$]{
\begin{minipage}[t]{0.32\linewidth}
\centering
\includegraphics[width=2.4in]{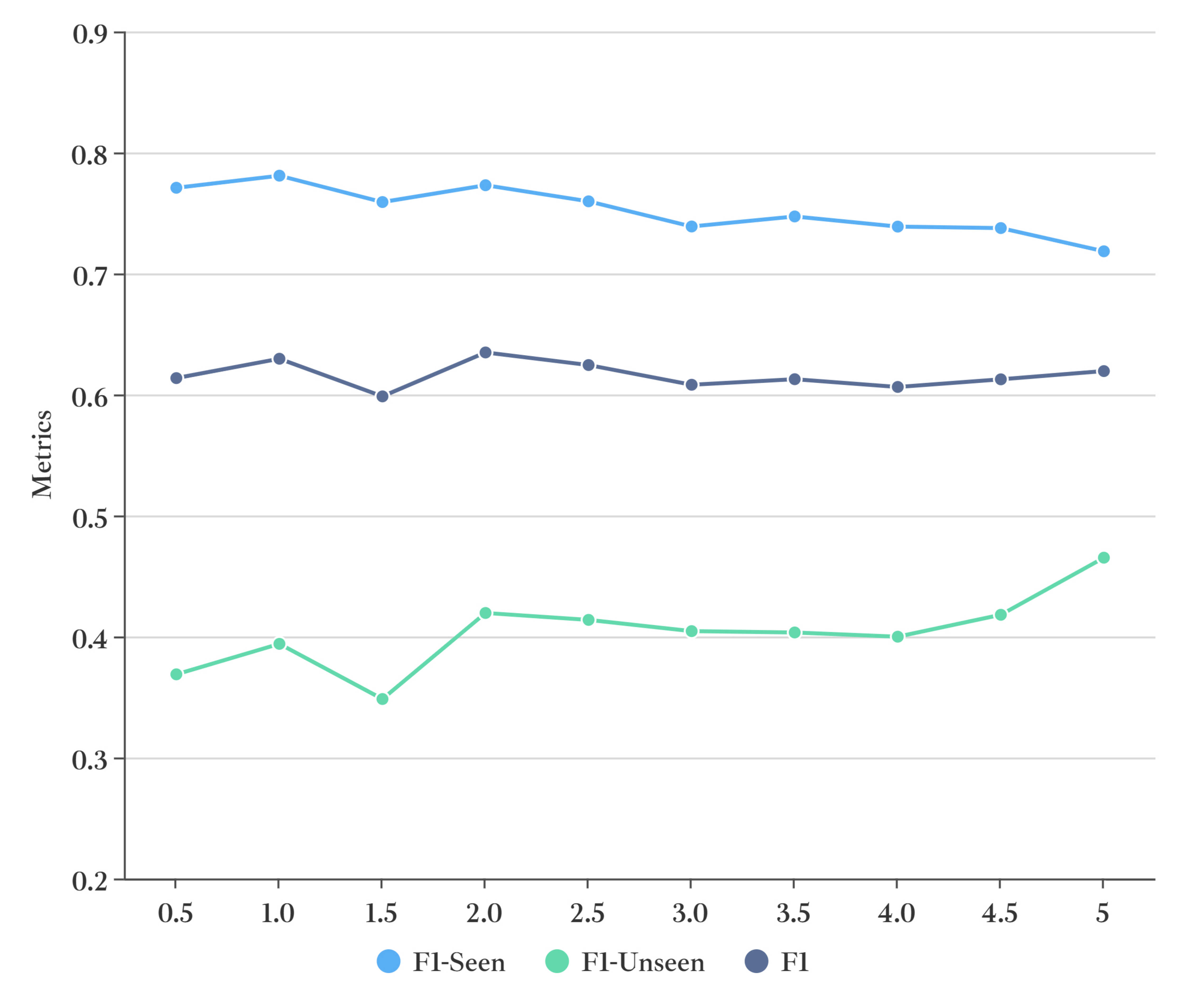}
\end{minipage}%
}%
\subfigure[$m_2$]{
\begin{minipage}[t]{0.32\linewidth}
\centering
\includegraphics[width=2.4in]{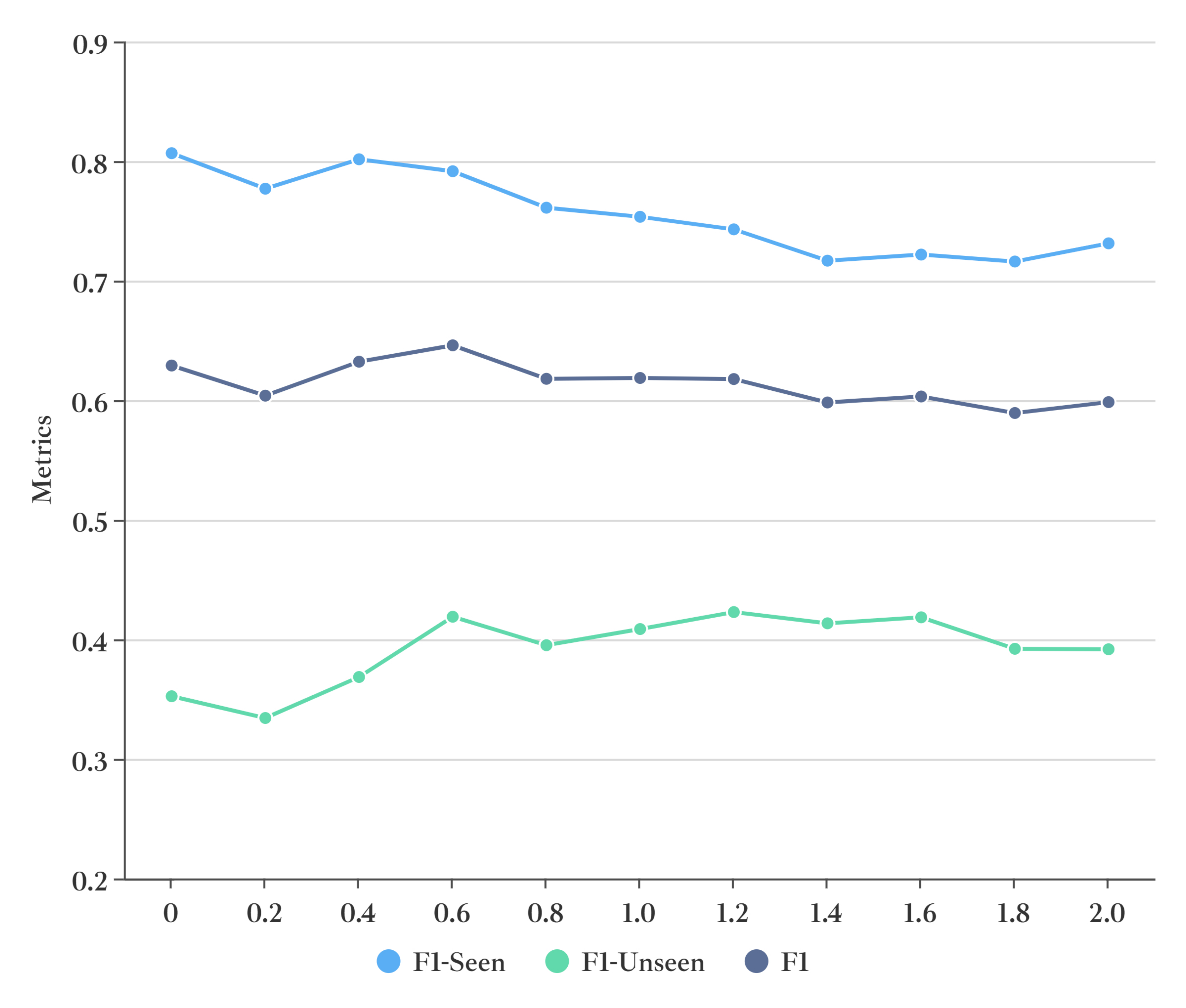}
\end{minipage}%
}%
\subfigure[$\lambda$]{
\begin{minipage}[t]{0.32\linewidth}
\centering
\includegraphics[width=2.4in]{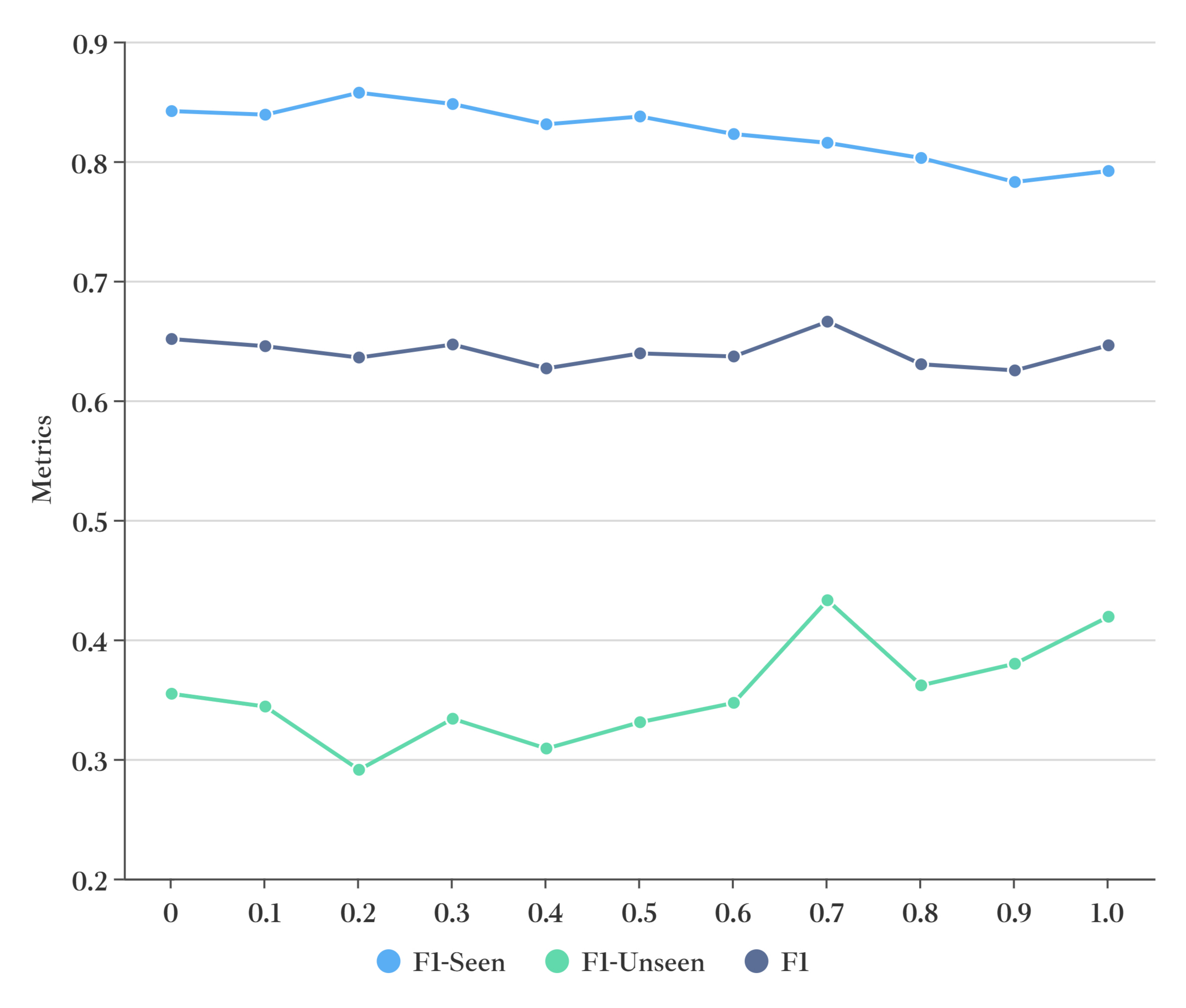}
\end{minipage}%
}%
\centering
\caption{Sensitivity analysis results of hyperparameters in F1-Seen, F1-Unseen and F1. }
\label{para}
\end{figure*}

\subsection{Ablation study}
We perform ablation experiments on the ACE 2005 and FewEvent datasets to examine the impact of removing certain components of the PBCT model. This aids in enhancing our comprehension regarding the significance of individual components within the model's overall performance. The results of ablation studies are shown in Table \ref{tab:ablation}.

To evaluate the influence of semantic initialization, we replace it with random initialization (denoted as -SemanInit.). This results in a decrease of F1 score by 6.15\% on the ACE 2005 dataset and by 0.68\% on the FewEvent dataset. Obviously, this difference indicates that semantic initialization is more crucial for training on small-scale datasets, as it can shorten the path of parameter optimization. Removing the masked contrastive learning  (i.e., -wo MaskContra.) leads to a slight overall decrease, except an increase in F1-Seen on the ACE 2005 dataset. This could be attributed to the model’s focus on refining seen event sample and type representations, thereby leading to higher F1 scores of seen event metrics. Furthermore, the model utilized for encoding seen event sample representations is identical to the one used for encoding unseen event sample representations, thereby indirectly optimizing the representations of unseen events. In addition, semantic initialization offers a high-quality type representation for unseen event types, ensuring that there is no significant decrease in the F1 scores for detecting unseen events. To evaluate the effectiveness of trigger-attentive sentinel, we replace it with the sum of contextual embedding vectors and trigger embedding vectors (denoted as -wo Atten.). Its inferior performance relative to PBCT suggests that effectively discriminating between different event types has been essential for enhancing performance.

\subsection{Trigger Saliency Analysis}

We leverage seen events from the ACE 2005 dataset to the unique characteristics between trigger-salient (TS) and context-salient (CS) types. Specifically, we compute the mean value of samples' $g_1$ within each event type $y_j\in \mathcal{T}_s$ on the training set $\mathcal{D}_s$ as trigger saliency $\overline{g}_j$: 
\begin{equation}\label{eq1}
\begin{aligned}
\overline{g}_j= \frac{1}{\lvert\mathcal{D}_s(y_j)\rvert-1}\sum_{i=0}^{\lvert\mathcal{D}_s(y_j)\rvert-1} g_1^{i} 
\\ j=0,1,\ldots, \lvert \mathcal{T}_s \rvert
\end{aligned}
\end{equation}
where $\mathcal{D}_s(y_j)$ represents seen event samples labeled as type $y_j$ in the training dataset.

We arrange $\overline{g}_j$ in descending order and divide event types into TS for the top half and CS for the bottom half, determined by the median. Afterward, we test the performance change in event detection of trigger-salient (TS) and context-salient (CS) types before and after removing masked contrastive learning and trigger-attentive sentinel, as shown in Fig. \ref{fig: CS}. Obviously, CS types exhibit a more substantial decline in performance compared to TS types, particularly in terms of precision. We believe this phenomenon arises from the model's post-removal tendency to focus on learning lexical patterns rather than developing a nuanced understanding of contextual cues. Consequently, classification tasks degrade into simplistic matching ones. The reliance of CS types on contexts makes them more sensitive to the removal of two components facilitating contextual understanding. 

\begin{figure*}[htbp]
\centering
\subfigure[SCCL]{
\begin{minipage}[t]{0.24\linewidth}
\centering
\includegraphics[width=1.7in]{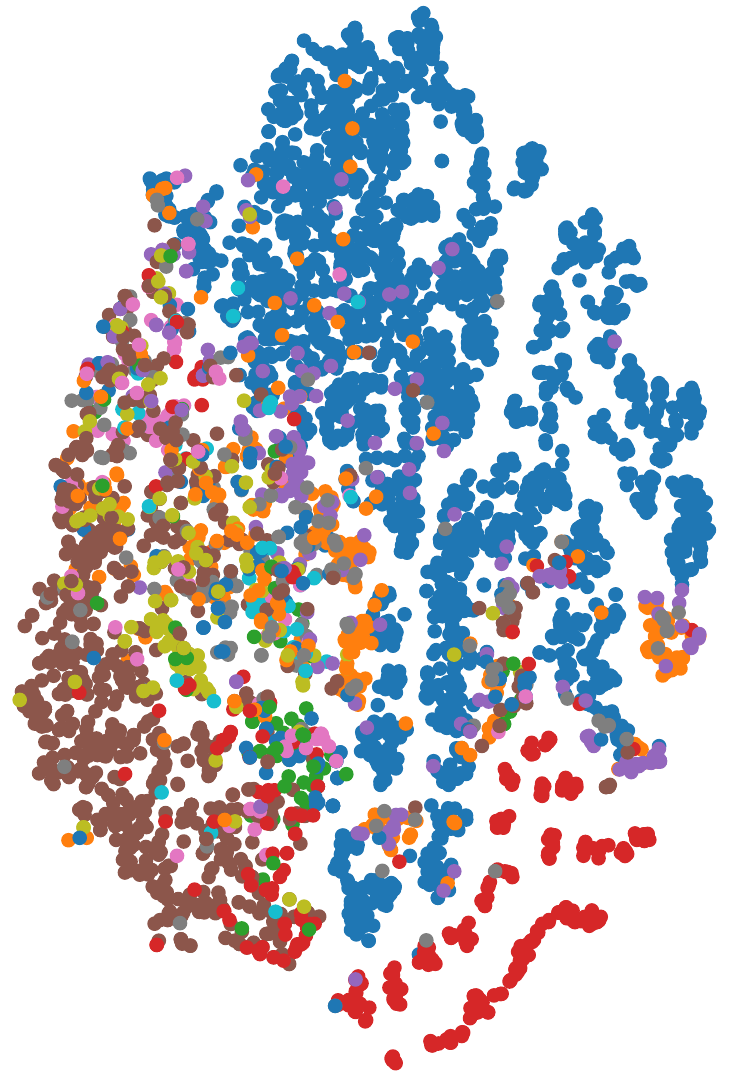}
\includegraphics[width=1.7in]{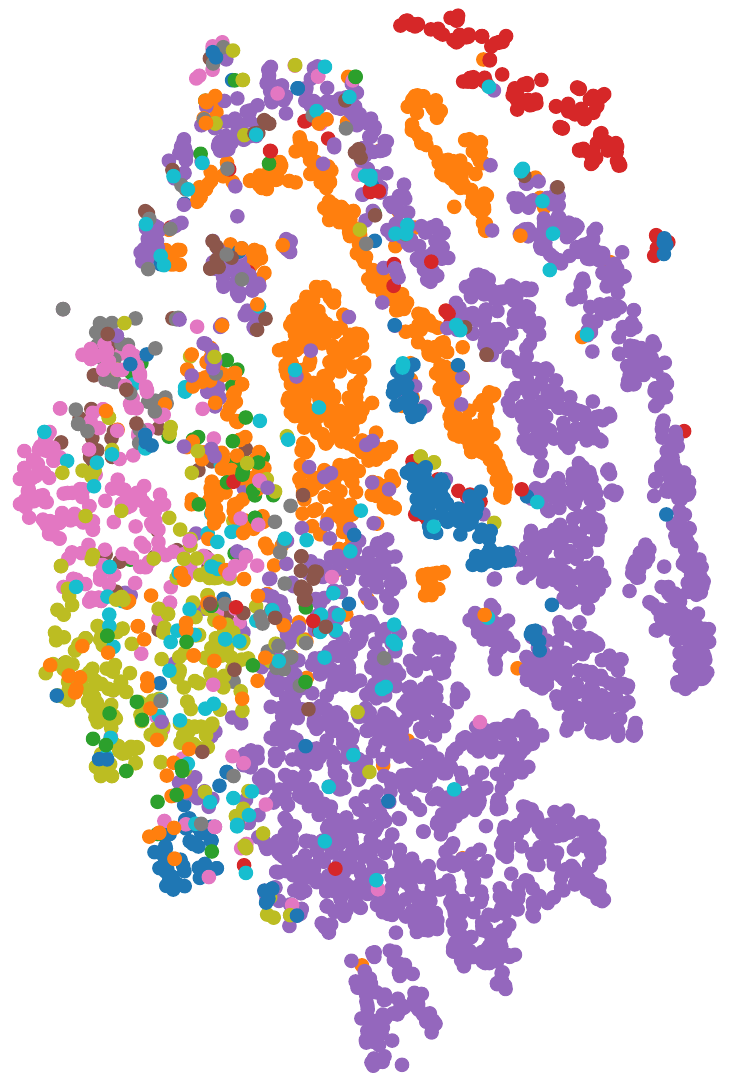}
\includegraphics[width=1.7in]{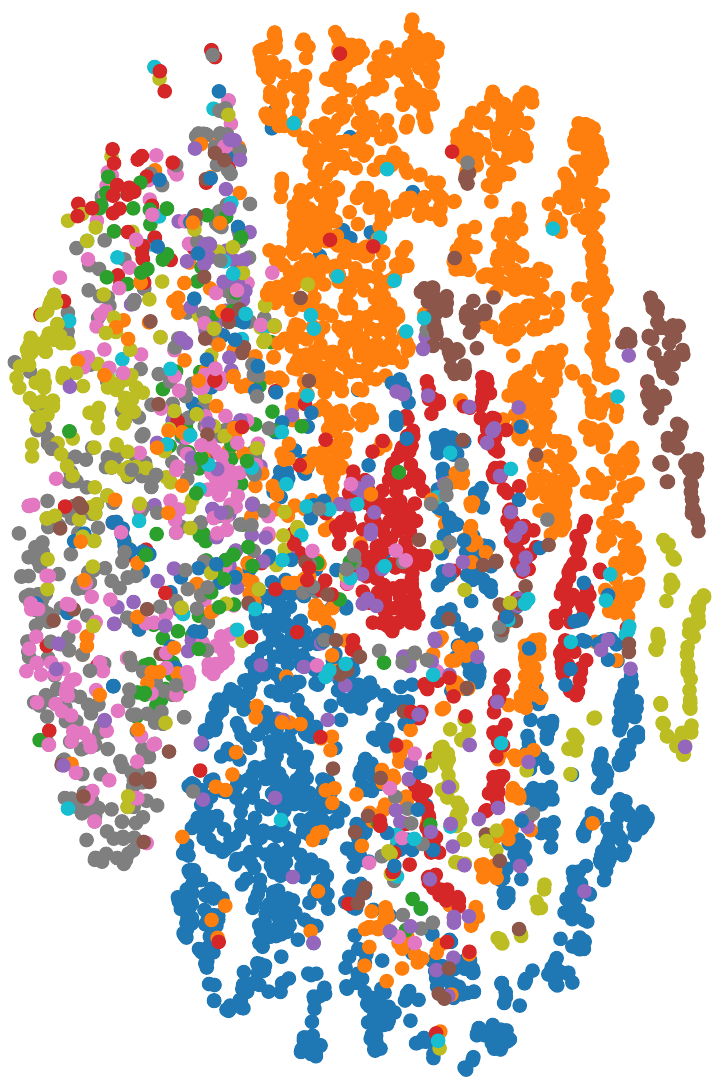}
\end{minipage}%
}%
\subfigure[SS-VQ-VAE]{
\begin{minipage}[t]{0.24\linewidth}
\centering
\includegraphics[width=1.7in]{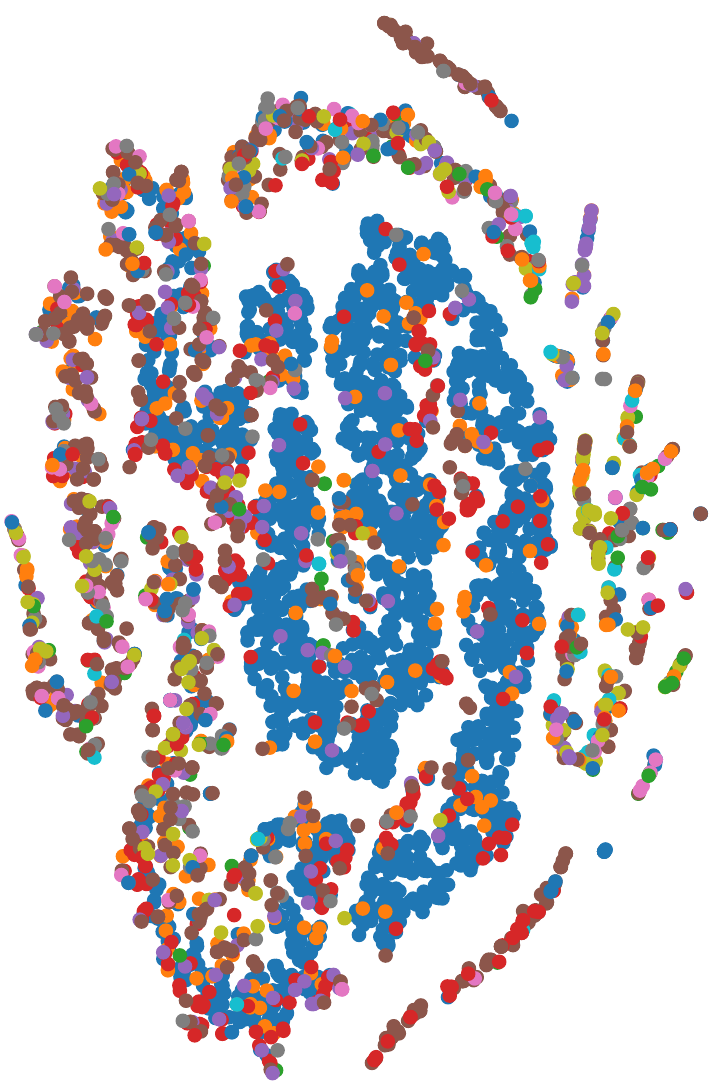}
\includegraphics[width=1.7in]{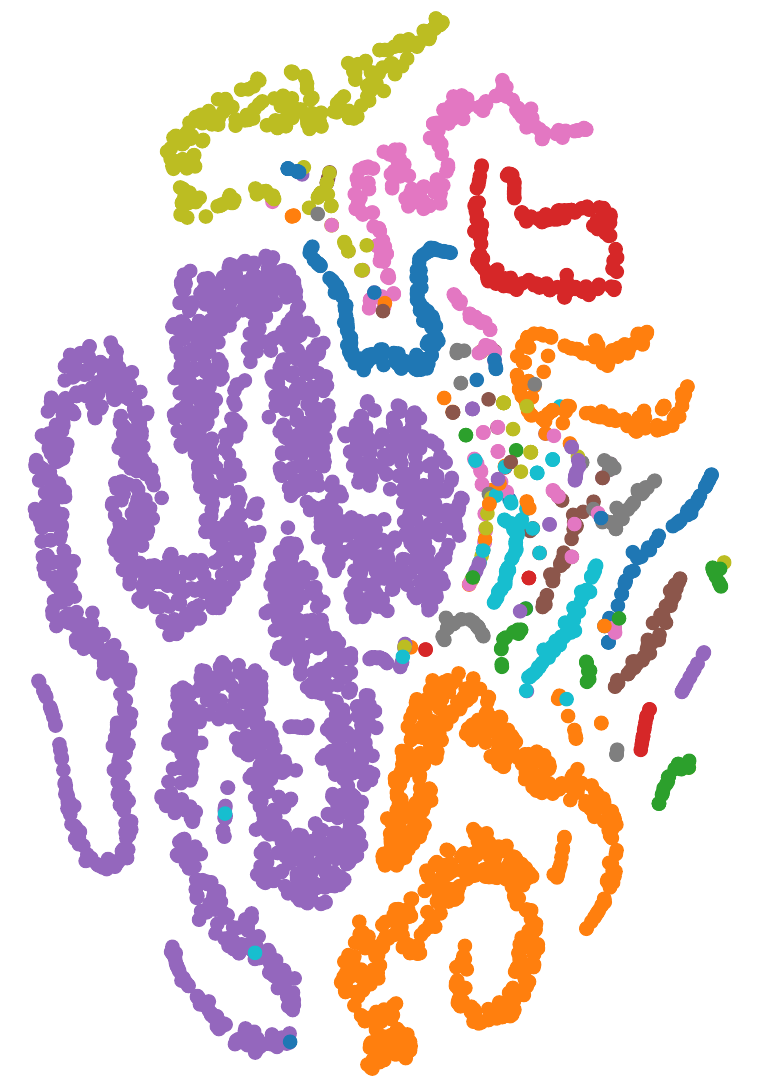}
\includegraphics[width=1.7in]{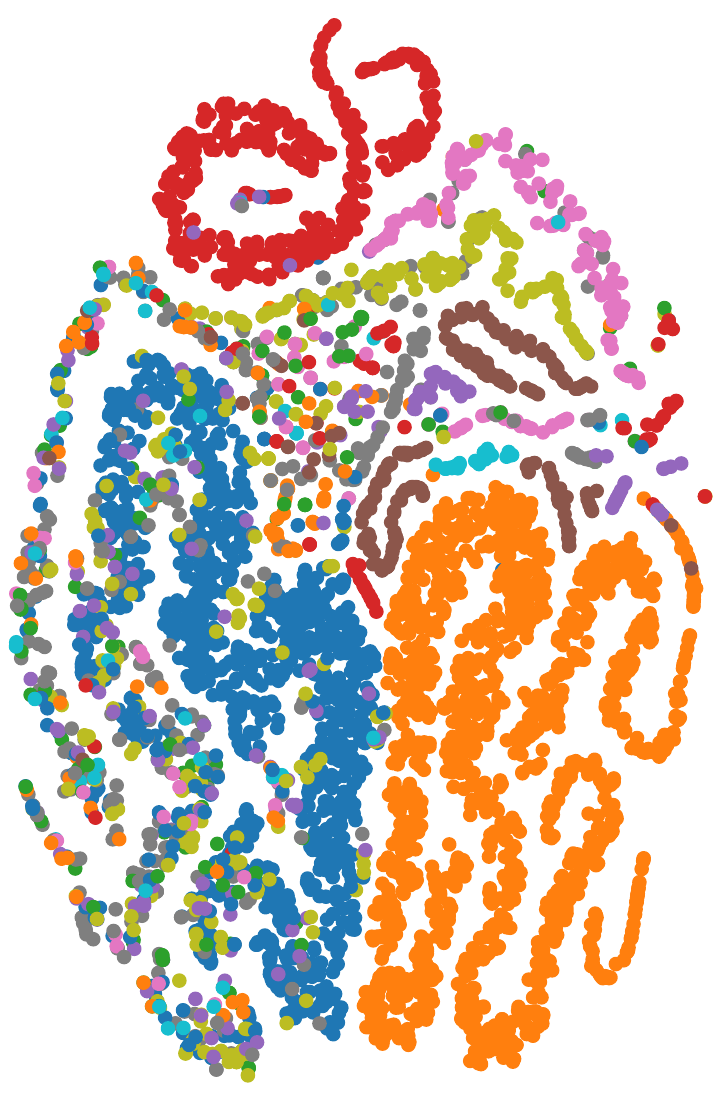}
\end{minipage}%
}%
\subfigure[BERT-MCL]{
\begin{minipage}[t]{0.24\linewidth}
\centering
\includegraphics[width=1.7in]{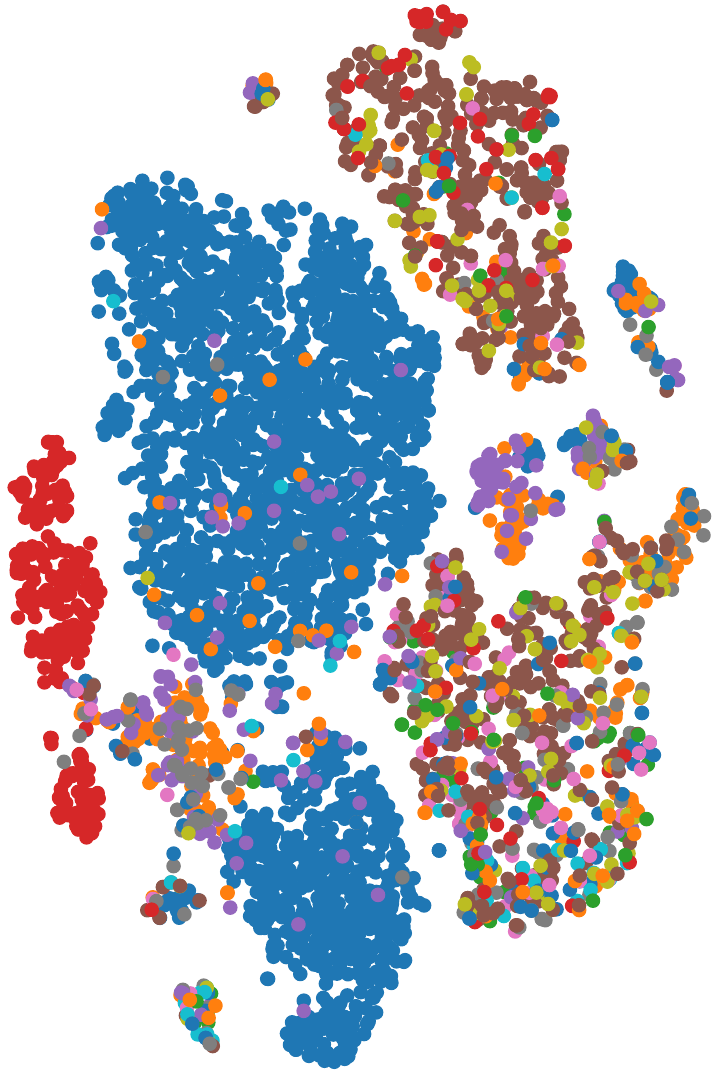}
\includegraphics[width=1.7in]{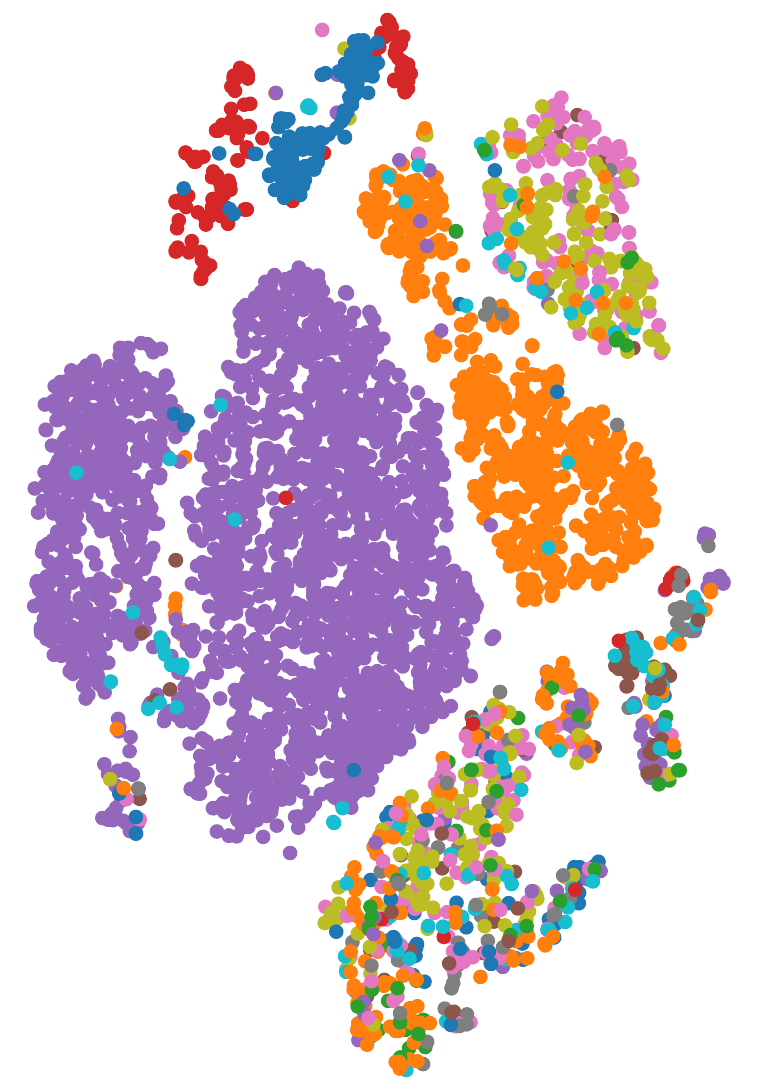}
\includegraphics[width=1.7in]{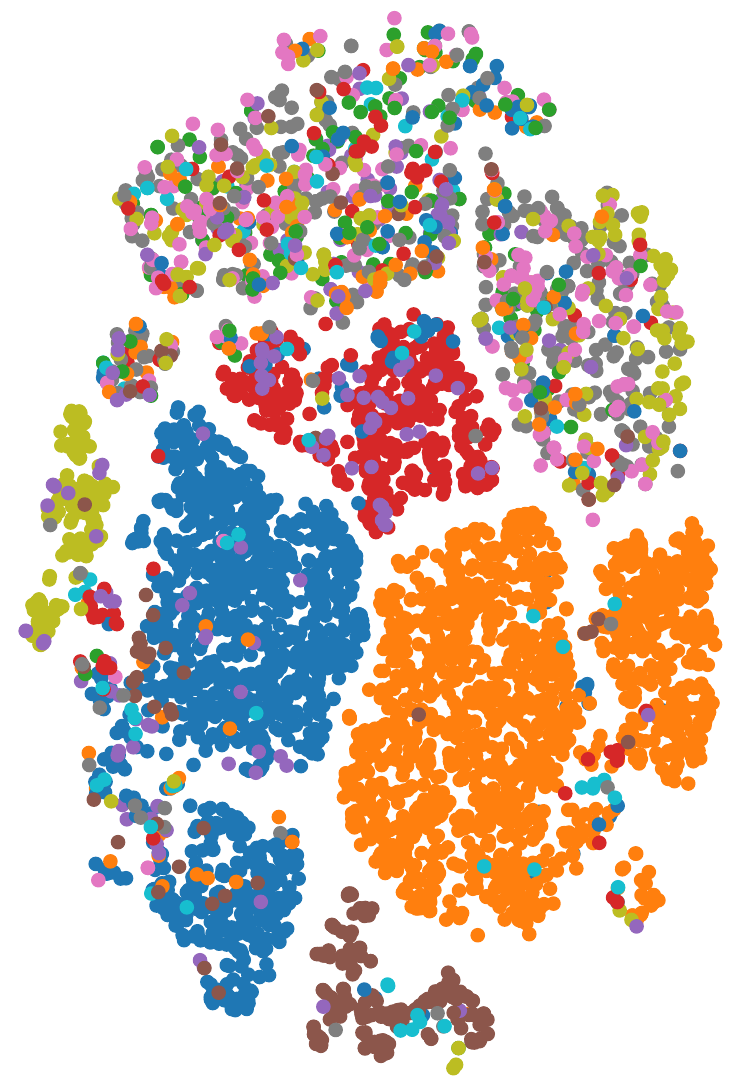}
\end{minipage}%
}%
\subfigure[PBCT]{
\begin{minipage}[t]{0.24\linewidth}
\centering
\includegraphics[width=1.7in]{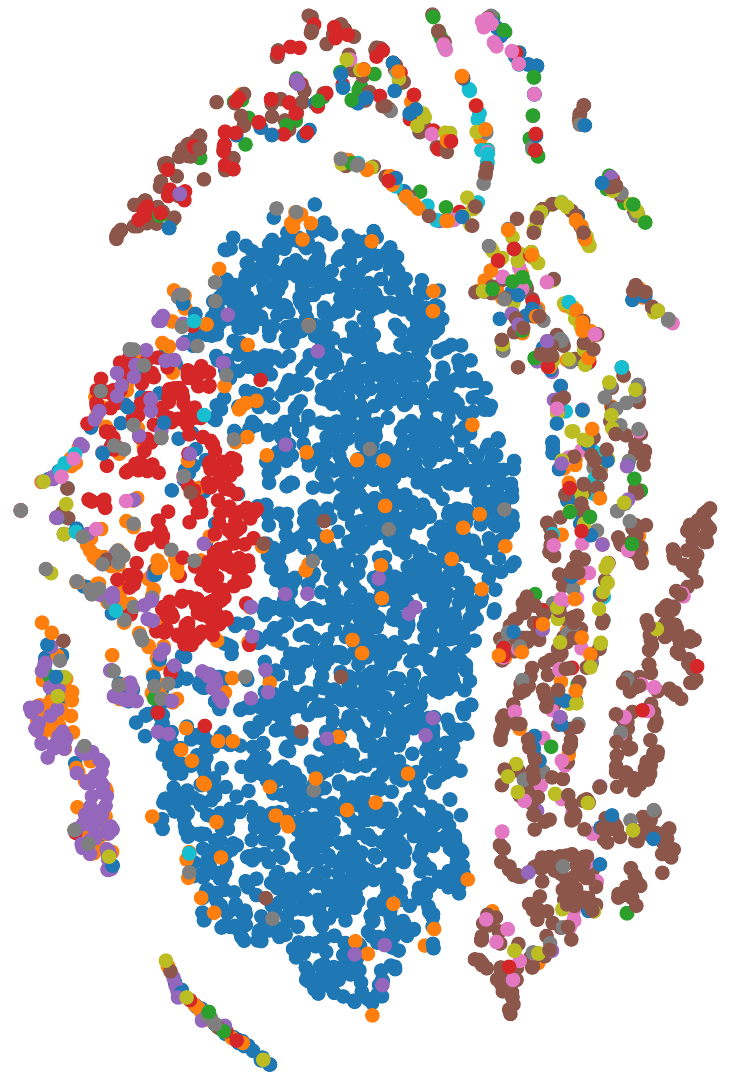}
\includegraphics[width=1.7in]{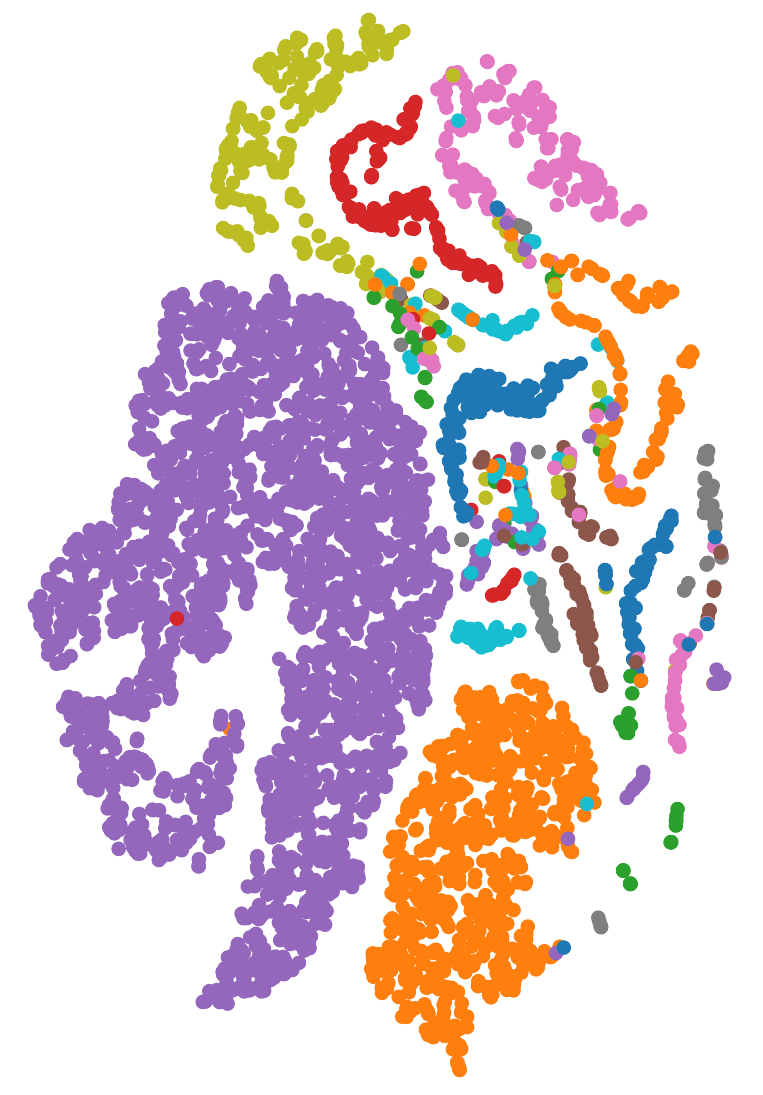}
\includegraphics[width=1.7in]{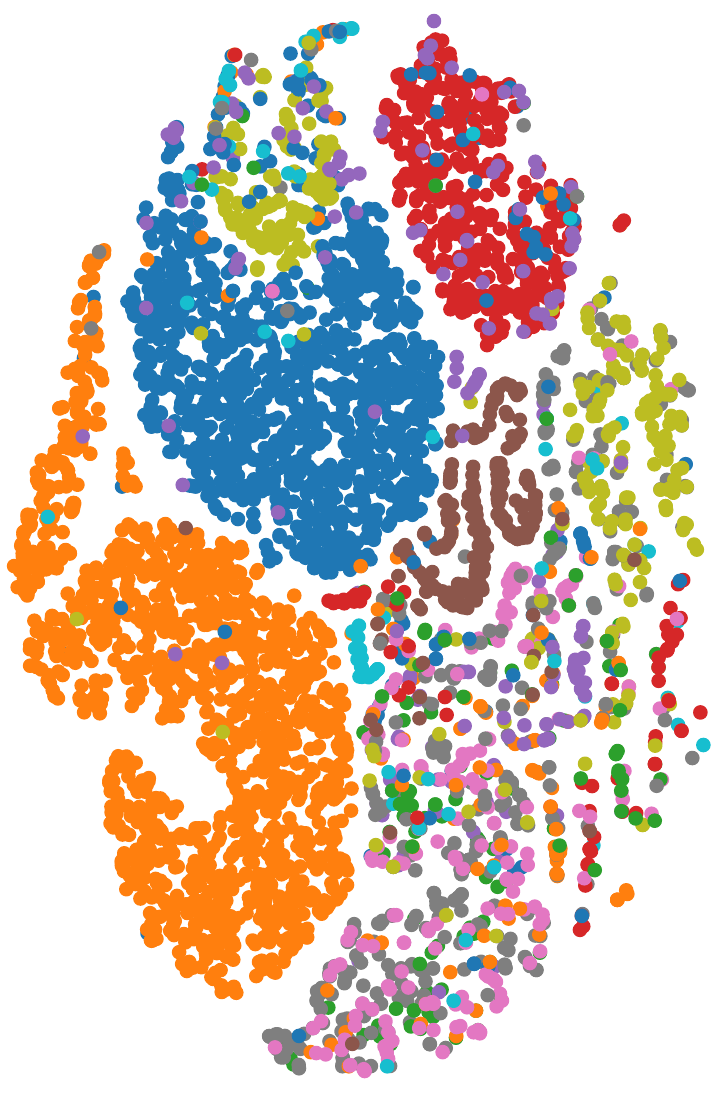}
\end{minipage}%
}%
\centering
\caption{Visualization of samples across various event types. From top to bottom are respectively the unseen event samples, seen event samples, and total event samples.}
\label{visual}
\end{figure*}

\subsection{Hyperparameter Analysis}
We investigate the impact of varying hyperparameters $m_1$, $m_2$ and $\lambda$ on PBCT performance using the ACE 2005 dataset, as shown in Fig. \ref{para}.

The hyperparameter $m_1$ regulates the scope of negative samples to be considered for pulling away. As depicted in Fig. \ref{para} (a), with higher values of $m_1$ than a certain boundary, F1-Seen decreases while F1-Unseen increases. This may be because the introduction of distant negative samples results in substantial contrastive loss, thereby impeding the thorough optimization of the cross-entropy loss. Therefore, with the comprehensive assessment consideration of the detecting performance of both event types, we employ the F1 score as the metric to select the optimal value for $m_1$, i.e., $m_1=2.0$.

Similar to $m_1$, the hyperparameter $m_2$ controls the range of masked samples that should be considered for pulling closer. Fig. \ref{para} (b) illustrates an initial ascent trend of F1 followed by a subsequent decline as $m_2$ increases, especially F1-Seen. We attribute this trend change to the distortion of representations of other events caused by forcibly pulling in masked samples from a greater distance.  $m=0.6$ is chosen as the optimal value as it yields the best  F1 scores for both F1 and F1-Unseen, along with competitive performance in F1-Seen.

The aim of tuning the hyperparameter $\lambda$ is to achieve a balance between the influences of supervised learning and contrastive learning. As shown in Fig. \ref{para} (c), there is a slight downward trend in the value of F1-Seen with an increase in the value of $\lambda$. This is evidently due to the reduced proportion of supervised loss, leading to insufficient training in seen event representations. Meanwhile, F1-Unseen demonstrates an overall upward trend, accompanied by significant fluctuations. $\lambda=0$ represents the absence of masked contrastive learning, and its relatively high F1-Unseen value is caused by effective supervised learning indirectly optimizing representations of unseen event samples. Therefore, we should select an appropriate $\lambda$ value that maximizes the benefits of contrastive learning without hindering supervised learning. Obviously, the optimal choice for $\lambda$ is 0.7. 

\subsection{Visualization Analysis}
For the visual observation of models’ representation learning, we employ t-SNE\footnote{https://scikit-learn.org/stable/modules/generated/sklearn.manifold.TSNE.html}  to visualize various event samples in the FewEvent dataset, with BERT-Base being used as the PLM, as shown in Fig. \ref{visual}. PBCT can achieve better representations of both unseen events and seen events, while SS-VQ-VAE performs poorly on unseen event types, and BERT-MCL on seen event types. This suggests that both supervised learning and contrastive learning play crucial roles in representation learning. By comparing the representations of seen and unseen events, we can discern that classifying unseen event samples is more intricate than classifying seen ones, with only a handful of high-sample-size types being accurately classified. Generally, the greater the number of samples within a type, the more effective the clustering.

\subsection{Case Study}
To validate the model's effectiveness in unfamiliar and specialized domains, we utilize ChatGPT to generate sentences describing the steps of PCB design and manufacturing. Subsequently, these sentences are formatted into a series of event detection samples and inputted into the PBCT model that has already been trained on the FewEvent dataset. Their input event mentions as well as the output recognized trigger words, and classified event types are presented in TABLE \ref{tab:example}. It's notable that despite the absence of PCB-specific knowledge in the FewEvent dataset, the model proficiently identifies trigger words and categorizes them under the \emph{Education.Education} label. However, the model is constrained by its ability to recognize only a single event, which is a focal point for our future efforts. 

\begin{table}[!t]
  \centering
  \caption{Event detection examples in unfamiliar domains using the trained PBCT model}\label{tab:example}
  {
    \begin{tabularx}{1.0\linewidth}
    { m{8cm} }
      \hline
      \emph{Event Mention}: It starts with creating a wiring layout using circuit design software.\\
      \emph{Trigger Word}: creating \\
       \emph{Event Type}: Education.Education\\
      \hline
       \emph{Event Mention}: It is followed by crafting a PCB schematic to represent the circuit connections.\\
       \emph{Trigger Word}: crafting \\
       \emph{Event Type}: Education.Education\\
      \hline
      \emph{Event Mention}: Next, the circuit diagram is transformed into an actual layout using PCB layout software. \\
       \emph{Trigger Word}: transformed \\
       \emph{Event Type}: Education.Education\\
      \hline
    \end{tabularx}
  }
\end{table}

\section{Conclusion and Future Work }
In this paper, we integrate RAG technology into the crowdsourcing framework to achieve automated crowdsourcing TD accurately and universally. By this means, TD tasks are reconfigured into event detection tasks. Therefore, we design a PBCT framework that incorporates prompt learning, a trigger-attentive sentinel and masked contrastive learning to address the reliance on external tools and the challenge of selectively handling different event types. The experimental results indicate our model performs comparably or even surpasses the state-of-the-art baselines in event detection for both seen and unseen events.

However, there are some limitations in establishing the workflow of decomposed subtasks for subsequent task recommendations. (1) The method is unable to address cases where multiple events are described within a single sentence, a scenario frequently encountered in real-world contexts. (2) The detected events (also known as subtasks) have temporal dependencies during execution. To allocate subtasks effectively, it is crucial to consider the temporal relationships between events. (3) It is assumed that relevant descriptions of task execution steps have already been retrieved prior to application before our models are applied, but in reality, the efficient and accurate retrieval of key content is of critical importance as it directly influences the performance of subsequent TD. Therefore, our research will focus on RAG-based crowdsourcing from three perspectives: multi-event detection, temporal relation extraction, and key content retrieval.

\bibliographystyle{IEEEtran}      
\bibliography{ref}

\vfill

\end{document}